\documentclass[nohyperref, dvipsnames]{article}

\usepackage{microtype}
\usepackage{graphicx}
\usepackage{subfigure}
\usepackage{booktabs} 

\usepackage{hyperref}
\usepackage{multirow}
\usepackage{subfloat}

\usepackage[accepted]{template/icml2023}

\usepackage{amsmath}
\usepackage{amssymb}
\usepackage{mathtools}
\usepackage{amsthm}

\usepackage{import}

\usepackage[capitalize,noabbrev]{cleveref}

\theoremstyle{plain}
\newtheorem{theorem}{Theorem}[section]
\newtheorem{proposition}[theorem]{Proposition}

\theoremstyle{definition}

\theoremstyle{remark}

\usepackage[textsize=tiny]{todonotes}

\newcommand{\V}{\mathcal{V}} 
\newcommand{\E}{\mathcal{E}} 
\newcommand{\mlp}{\mathrm{MLP}}
\newcommand{\RR}{\mathbb{R}}
\newcommand{\rw}{\mathbf{M}}
\newcommand{\spd}{\mathrm{SPD}}
\renewcommand{\P}{\mathbf{P}}
\def\ourmethod {GRIT }

\icmltitlerunning{Graph Inductive Biases in Transformers without Message Passing}

\begin{document}

\twocolumn[
\icmltitle{Graph Inductive Biases in Transformers without Message Passing}



\icmlsetsymbol{equal}{*}
\icmlsetsymbol{ills2}{8}

\begin{icmlauthorlist}
\icmlauthor{Liheng Ma}{equal,mcgill,ills2}
\icmlauthor{Chen Lin}{equal,oxford}
\icmlauthor{Derek Lim}{mit}
\icmlauthor{Adriana Romero-Soriano}{metaai,mcgill,mila,cifar}
\icmlauthor{Puneet K. Dokania}{oxford,fiveai}

\icmlauthor{Mark Coates}{mcgill,ills}
\icmlauthor{Philip H.S. Torr}{oxford}
\icmlauthor{Ser-Nam Lim}{metaai}
\end{icmlauthorlist}

\icmlaffiliation{mcgill}{McGill University}
\icmlaffiliation{oxford}{Department of Engineering Science, University of Oxford}
\icmlaffiliation{mit}{CSAIL, Massachusetts Institute of Technology}
\icmlaffiliation{fiveai}{FiveAI}
\icmlaffiliation{metaai}{MetaAI}
\icmlaffiliation{mila}{Mila - Quebec AI Institute} 
\icmlaffiliation{ills}{International Laboratory on Learning Systems (ILLS)} 
\icmlaffiliation{cifar}{Canada CIFAR AI Chair} 

\icmlcorrespondingauthor{Liheng Ma}{liheng.ma@mail.mcgill.ca}
\icmlcorrespondingauthor{Chen Lin}{chen.lin@eng.ox.ac.uk}

\icmlkeywords{Graph Transformers, Message Passing, Inductive Bias}

\vskip 0.3in
]



\printAffiliationsAndNotice{\icmlEqualContribution} 

\begin{abstract}

Transformers for graph data are increasingly widely studied and successful in numerous learning tasks.  
Graph inductive biases are crucial for Graph Transformers, and previous works incorporate them using message-passing modules and/or positional encodings.
However, Graph Transformers that use message-passing inherit known issues of message-passing, and differ significantly from Transformers used in other domains, thus making transfer of research advances more difficult. 
On the other hand, Graph Transformers without message-passing often perform poorly on smaller datasets, where inductive biases are more important.
To bridge this gap, we propose the Graph Inductive bias Transformer (GRIT) --- a new Graph Transformer that incorporates graph inductive biases without using message passing. GRIT is based on several architectural changes that are each theoretically and empirically justified, including: learned relative positional encodings initialized with random walk probabilities, a flexible attention mechanism that updates node and node-pair representations, and injection of degree information in each layer. We prove that GRIT is expressive --- it can express shortest path distances and various graph propagation matrices. GRIT achieves state-of-the-art empirical performance across a variety of graph datasets,
thus showing the power that Graph Transformers without message-passing can deliver.







\end{abstract}

\section{Introduction}

Following the success of Transformers~\citep{vaswani2017AttentionAllYou} in different modalities like natural language processing (NLP)~\citep{vaswani2017AttentionAllYou} and computer vision~\citep{dosovitskiy2020ImageWorth16x16}, 
developing Transformers for graph data has attracted much interest ~\cite{dwivedi2021GeneralizationTransformerNetworks, kreuzer2021RethinkingGraphTransformers, ying2021TransformersReallyPerform, chen2022StructureAwareTransformerGraph, hussain2022GlobalSelfAttentionReplacement, rampasek2022RecipeGeneralPowerful, zhang2023rethinking}. 
A major motivation of Graph Transformers is to alleviate certain known
limitations of (local) Message-Passing Graph Neural Networks (MPNNs)~\cite{ gilmer2017NeuralMessagePassing}, such as over-smoothing~\cite{li2018DeeperInsightsGraph, oono2020graph}, over-squashing~\cite{alon2020BottleneckGraphNeural, topping2022UnderstandingOversquashingBottlenecks}, and expressive power limitations~\cite{xu2019HowPowerfulAre, loukas2020WhatGraphNeural, morris2019WeisfeilerLemanGo}.




However, it is known that transformers generally \textit{lack strong inductive biases}~\cite{dosovitskiy2020ImageWorth16x16}.
In the graph domain, Graph Transformers aggregate information based on a learned attention matrix, which enjoys a high degree of freedom; in contrast, MPNNs explicitly aggregate the information according to the exact topology of the input graph.
This comes with at least two consequences.
First, Graph Transformers may be prone to over-fitting, and thus they often fail to outperform MPNNs in limited data settings.
Second, learning meaningful attention scores typically requires capturing important positional or structural relationships between nodes, 
and strong positional encodings are challenging to design since the structure and symmetries of graph data are fundamentally different from that of other (Euclidean) domains ~\cite{vaswani2017AttentionAllYou, bronstein2021GeometricDeepLearning}.
For instance, there is no ordering or canonical coordinate system for nodes in a graph, whereas words in a sentence have a sequence structure, and pixels in an image have a grid structure.

To incorporate graph inductive biases, many of the best-performing Graph Transformers explicitly integrate local message-passing mechanisms.
For instance, some works incorporate sparse attention on local neighborhoods \cite{dwivedi2021GeneralizationTransformerNetworks, kreuzer2021RethinkingGraphTransformers}, 
or integrate various other types of MPNN modules into their models~\cite{chen2022StructureAwareTransformerGraph, rampasek2022RecipeGeneralPowerful}.
Thus, such Graph Transformers may at least partially inherit some of the limitations of MPNNs.
Moreover, message-passing modules within Graph Transformers make these models significantly different from the Transformers used in other domains; thus, it becomes more difficult to transfer some of the large quantities of Transformer research from other domains into the graph domain.
This is exacerbated by the fact that integrating message-passing
adds complexity to the design space of the underlying model, necessitating more architectural decisions and extra effort for hyperparameter tuning \cite{masters2022GPSOptimisedHybrid}.

The trade-off between the limitations of message-passing and the importance of graph inductive biases can be seen in the empirical performance of models on competitive benchmarks. For instance, let us consider the results for two popular molecular graph regression benchmarks as of May 2023. On the small ZINC dataset (12,000 graphs)~\citep{dwivedi2020BenchmarkingGraphNeural}, GNNs that rely on message-passing take up the top spots on the leaderboards.\footnote{\url{https://paperswithcode.com/sota/graph-regression-on-zinc-500k} at the time of submission.} For the large PCQM4MV2 dataset (about 3,700,000 graphs)~\citep{hu2021ogblsc}, Graph Transformers take up the top spots.\footnote{\url{https://ogb.stanford.edu/docs/lsc/leaderboards/}}

In this work, we introduce the \underline{Gr}aph \underline{I}nductive bias \underline{T}ransformer (GRIT), a Graph Transformer that incorporates useful graph inductive biases without explicit message-passing modules. Our model is based on three design choices that integrate graph inductive biases, each of which is theoretically justified: (a) we use a learned relative positional encoding initialized with Relative Random Walk Probabilities (RRWP); this learned positional encoding can provably express shortest path distances~\citep{ying2021TransformersReallyPerform} and general classes of message-passing propagations~\citep{gasteiger2019DiffusionImprovesGraph, zhao2021AdaptiveDiffusionGraph, xu2019HowPowerfulAre}. (b) we develop an attention mechanism that jointly updates both node representations and node-pair representations, and can thus learn to update the RRWP positional encodings; a distance based Weisfeiler Leman test~\citep{zhang2023rethinking} shows that certain Graph Transformers with RRWP are strictly stronger than Graph Transformer with shortest path distances like Graphormer~\citep{ying2021TransformersReallyPerform}. (c) we inject degree information into our Transformer update using degree scalers, with batch normalization replacing the standard layer normalization; replacing the layer normalization is provably required to maintain the degree information.


Along with theoretical justification, we provide ample empirical evidence to demonstrate the effectiveness of our design choices. GRIT achieves state-of-the-art empirical performance across a variety of graph learning benchmarks, both small and large-scale. In particular, we bridge the performance gap in which message-passing-based methods do not perform as well in large datasets, and non-message-passing Transformers do not perform as well in small datasets. Ablations and synthetic experiments further justify our design decisions, showing that GRIT indeed integrates graph inductive biases into Transformers in an effective manner.\footnote{The code and models are publicly available at \url{https://github.com/LiamMa/GRIT}.}

\section{Related Work}

\paragraph{Transformers for Euclidean Domains}

Transformers have achieved ground-breaking successes in various domains, including but not limited to natural language processing~\cite{vaswani2017AttentionAllYou, devlin2019BERTPretrainingDeep, dai2019TransformerXLAttentiveLanguage} and computer vision~\cite{dosovitskiy2020ImageWorth16x16, liu2021SwinTransformerHierarchical}.
As can be seen in both domains, Transformers often suffer from a lack of inductive bias compared to Recurrent Neural Networks (RNNs) and Convolutional Neural Networks (CNNs), and typically require a large amount of training data in order to perform well.
Recent studies have found that introducing more inductive biases can effectively improve the performance of Transformers~\cite{liu2021SwinTransformerHierarchical, park2021HowVisionTransformers}

\paragraph{Positional Encoding and Structural Encoding for Graphs}

In the graph domain, positional encodings (PE) and structural encodings (SE)~\citep{srinivasan2019EquivalencePositionalNode} have been studied for enhancing both Message-Passing Graph Neural Networks (MPNNs) and Graph Transformers
~\cite{you2019PositionawareGraphNeural,ma2021graph,li2020DistanceEncodingDesign,zhang2021EigenGNNGraphStructure, dwivedi2020BenchmarkingGraphNeural, loukas2020WhatGraphNeural, dwivedi2021GraphNeuralNetworks, lim2022SignBasisInvariant, wang2022EquivariantStablePositional}.
Such positional or structural encodings capture various types of graph features, such as shortest-path distances~\cite{li2020DistanceEncodingDesign}, identity-awareness~\cite{you2021IdentityawareGraphNeural}, and spectral information~\cite{dwivedi2020BenchmarkingGraphNeural}.

Many so-called positional or structural encodings contain both
positional and structural information ~\cite{dwivedi2020BenchmarkingGraphNeural, srinivasan2019EquivalencePositionalNode, rampasek2022RecipeGeneralPowerful}; 
thus, researchers use the terms positional encoding and structural encoding interchangeably in related literature. In this work, we use ``positional encodings'' as an umbrella term.

\paragraph{Graph Transformers with Message-Passing}

\citet{dwivedi2021GeneralizationTransformerNetworks} is one of the early works that introduce a Transformer architecture to graph domains with Laplacian Positional Encoding (LapPE).
However, the global attention version of it performs poorly compared to the message-passing-based sparse attention version.
\citet{kreuzer2021RethinkingGraphTransformers} propose the SAN model, which uses both sparse and global attention mechanisms at each layer, 
and introduces an extra transformer to encode LapPE. 
SignNet~\citep{lim2022SignBasisInvariant} is a specialized symmetry-invariant encoder for LapPE that uses message-passing modules to process Laplacian eigenvectors~\citep{lim2022SignBasisInvariant, rampasek2022RecipeGeneralPowerful}.

Several Graph Transformers~\cite{chen2022StructureAwareTransformerGraph, rampasek2022RecipeGeneralPowerful} use random walk structural encodings (RWSE)~\cite{dwivedi2021GraphNeuralNetworks}. 
Due in part to the fact that RWSE focuses on structural information and does not encode as much positional information,
these Graph Transformers each integrate message-passing modules.

\paragraph{Graph Transformers without Message-Passing}

In contrast to the aforementioned works, a number of Graph Transformers without (local) message-passing have been proposed. 
Among them, a series of works propose to use relative positional encodings for each pair of nodes, such as shortest-path distance positional encodings (SPDPE)~\cite{ying2021TransformersReallyPerform, park2022GRPERelativePositional, luo2022your}.
Generalized-Distance Transformers~\cite{zhang2023rethinking}  introduce other distances on graphs (e.g., resistance distances) as relative positional encodings.  
These works are among the most related to our method and typically perform well on large datasets such as the PCQM4Mv2 dataset from the OGB Large Scale Challenge~\cite{hu2021ogblsc}.
However, they perform worse on smaller datasets (e.g., ZINC~\cite{dwivedi2020BenchmarkingGraphNeural}) compared to hybrid Graph Transformers and MPNNs~\cite{rampasek2022RecipeGeneralPowerful}. 
Based on a synthetic experiment (Sec.~\ref{sec:attn2mp}), we give evidence that due to the lack of sufficient inductive bias, they are not as capable in enabling attention mechanisms to perform local message-passing when necessary.

There are also several other Graph Transformers without message-passing proposed in recent years:
TokenGT~\cite{kim2022PureTransformersAre} proposes a Transformer that views both nodes and edges as tokens, and uses LapPE or  orthogonal random features; Relational Transformer~\citep{diao2022relational} also proposes to update both node and edge tokens;
EGT~\cite{hussain2022GlobalSelfAttentionReplacement} proposes to use an SVD-based PE instead of LapPE for directed graphs;
\cite{mialon2021GraphiTEncodingGraph, feldman2022weisfeiler} introduce positional encodings based on heat kernels and other graph kernels.

\begin{figure*}[ht]
\newcommand{\wth}{.20}
\centering
{
\hspace{8pt} \includegraphics[width=\wth\textwidth]{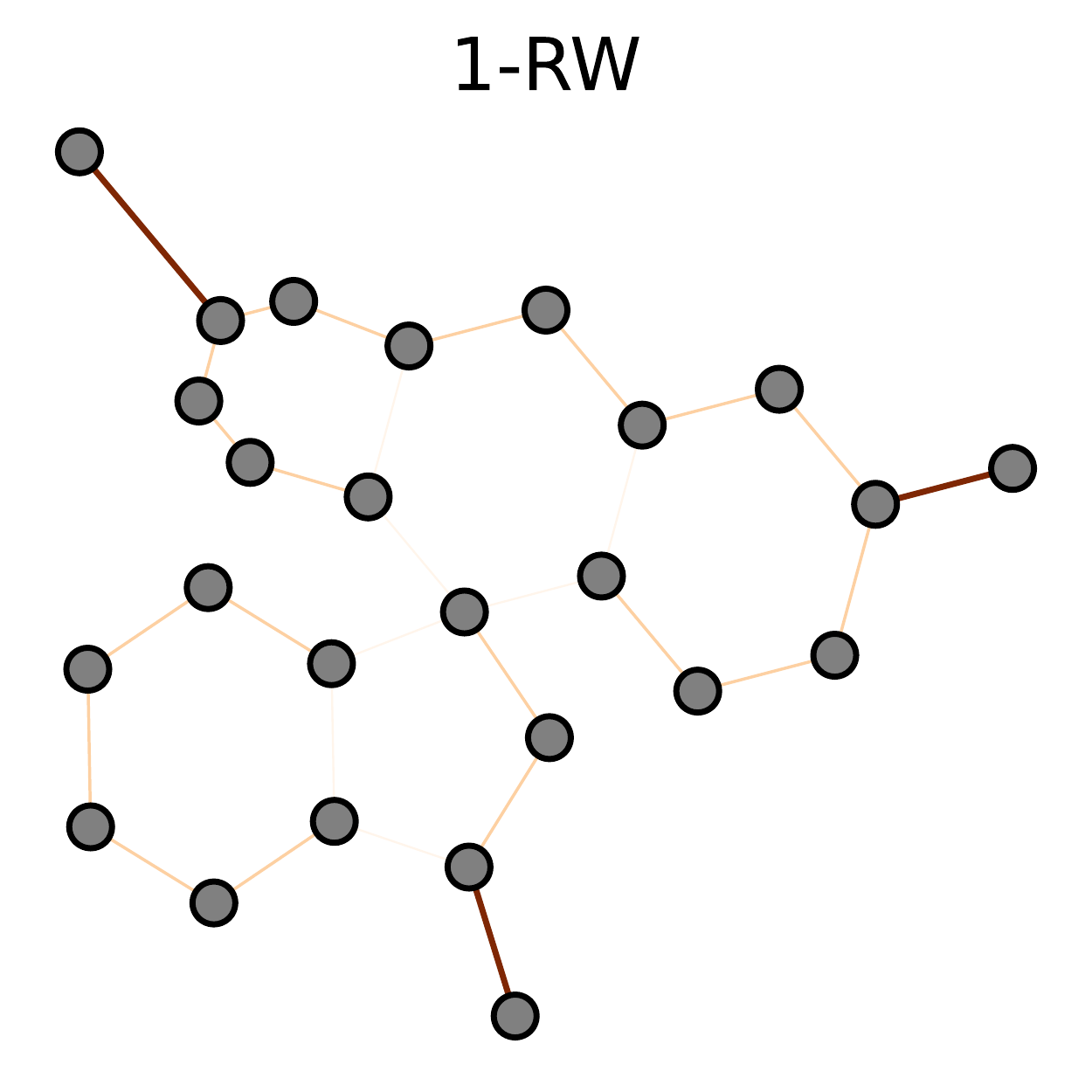}
\includegraphics[width=\wth\textwidth]{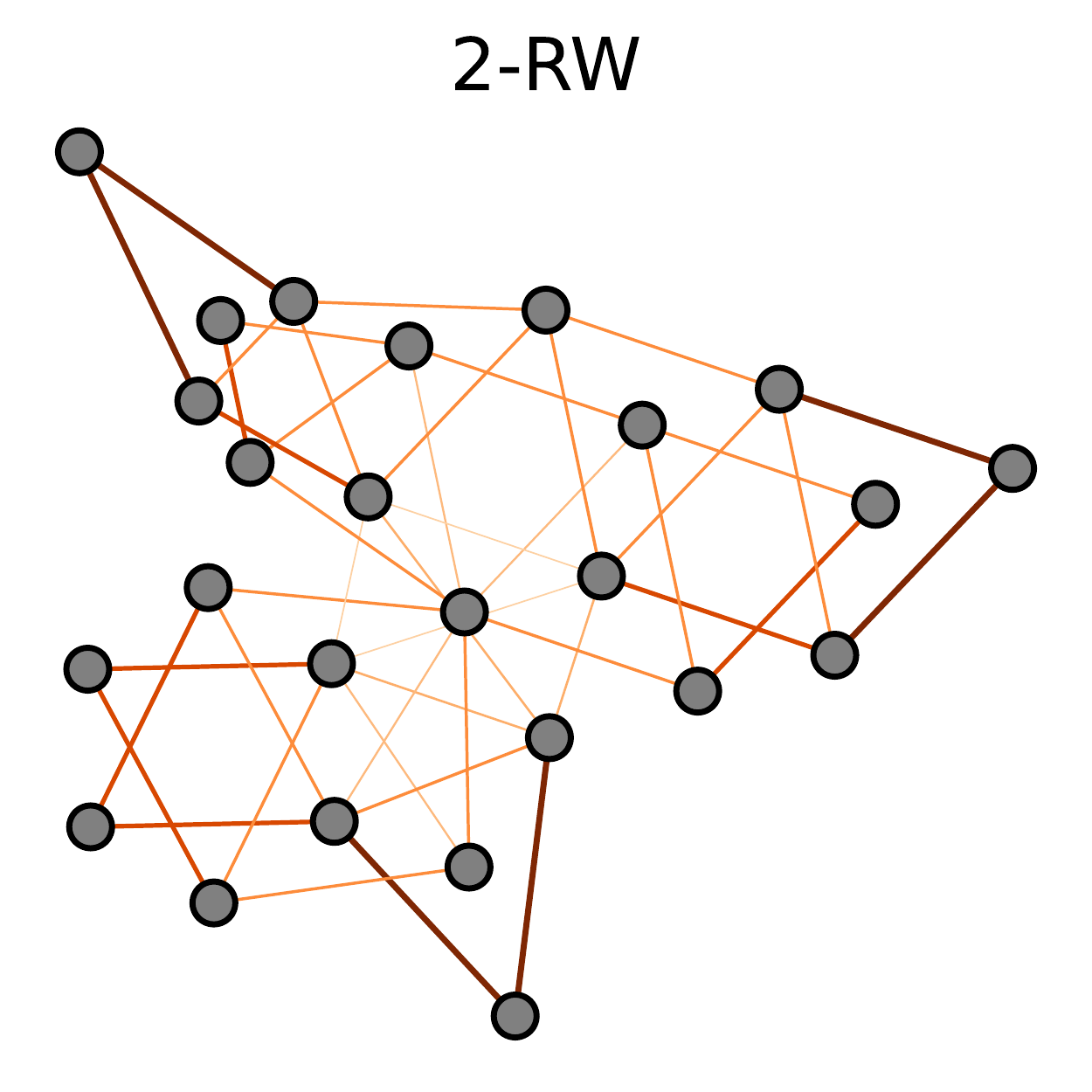}
\includegraphics[width=\wth\textwidth]{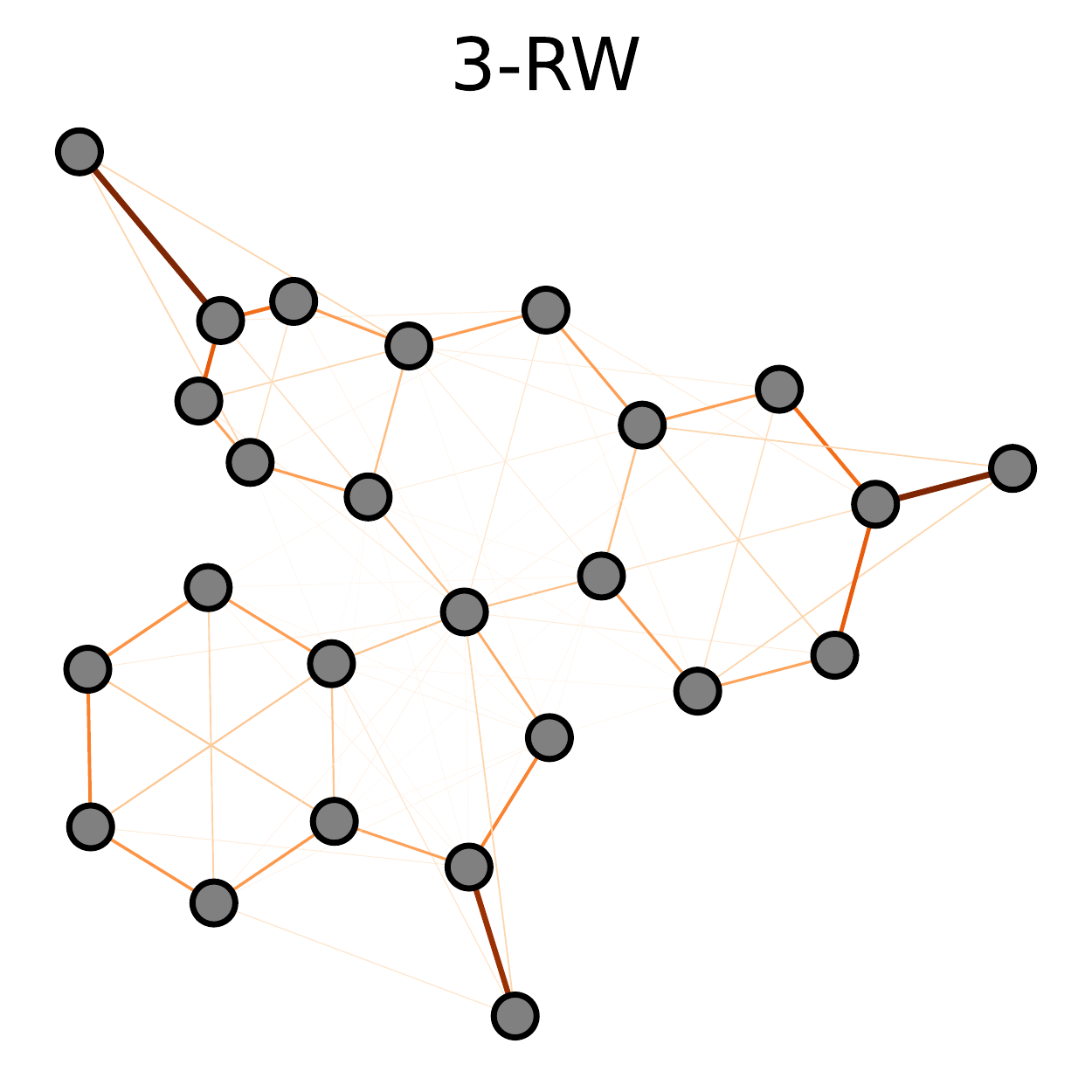}
\includegraphics[width=\wth\textwidth]{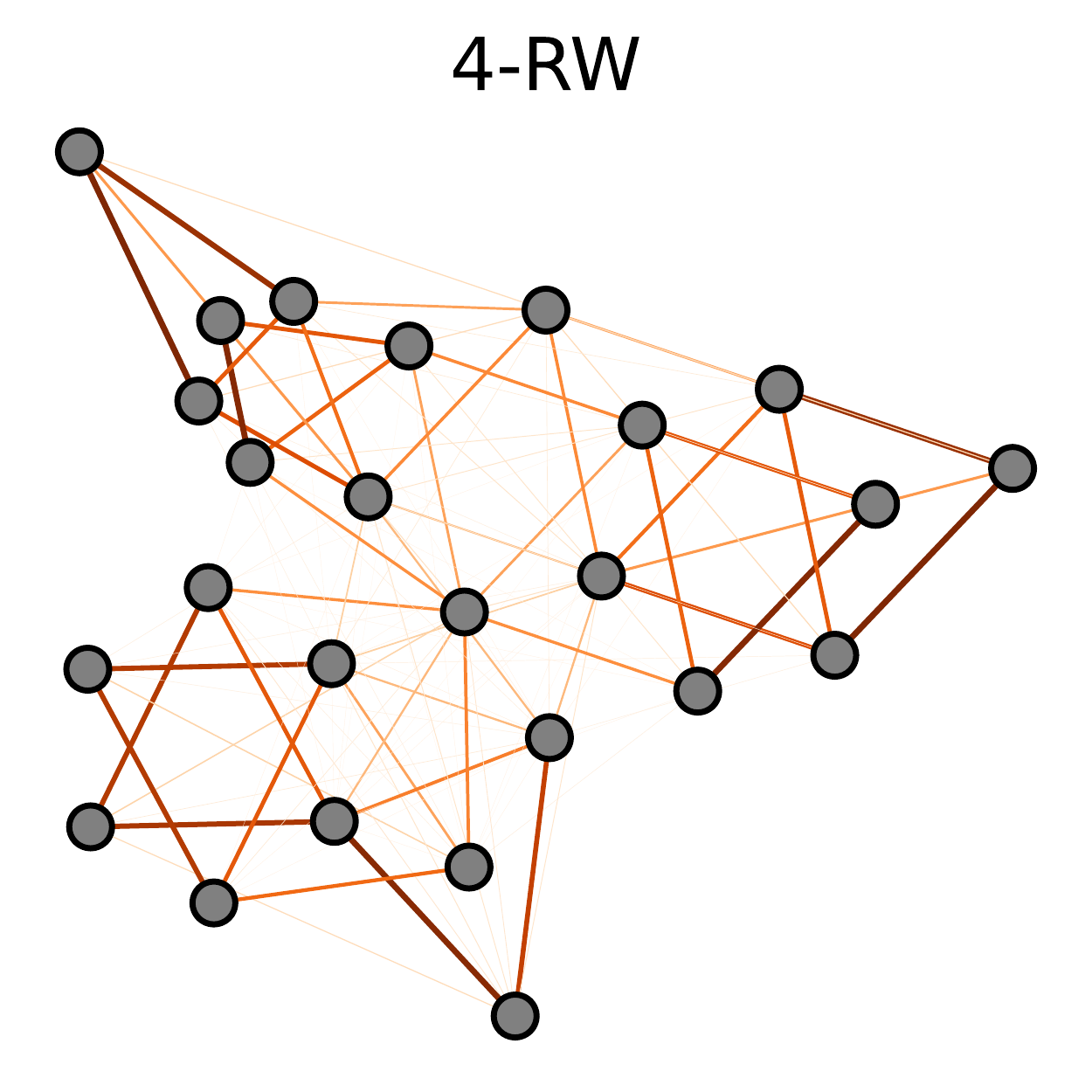}
\\[10pt]
}
\caption{RRWP visualization for the fluorescein molecule, up to the 4th power. Thicker and darker edges indicate higher edge weight. Probabilities for longer random walks reveal higher-order structures (e.g., the cliques evident in 3-RW and the star patterns in 4-RW).}
\label{fig:rw_viz_mol}
\end{figure*}

\begin{figure*}[ht]
\newcommand{\wth}{.20}
\centering
{
\hspace{8pt} \includegraphics[width=\wth\textwidth]{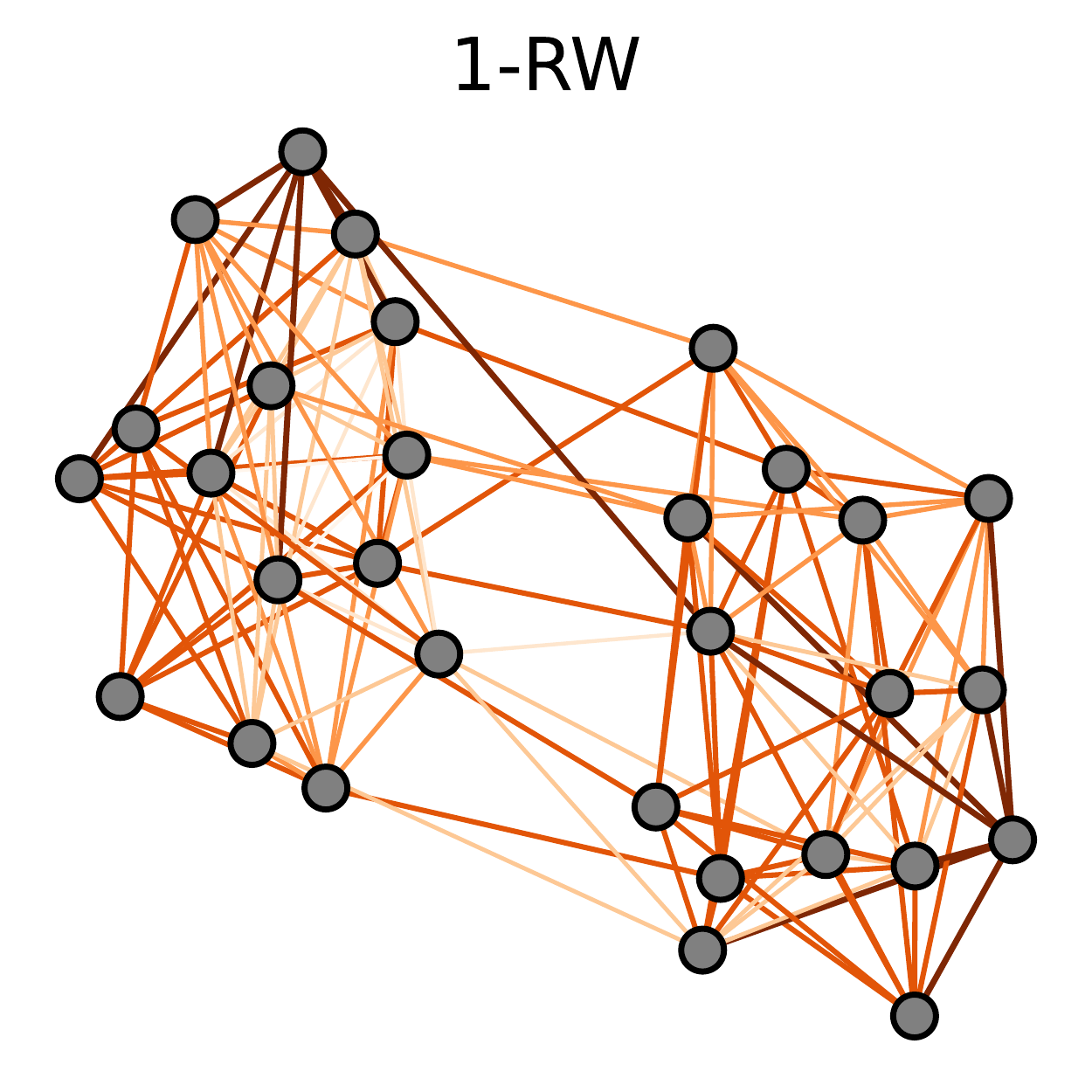}
\includegraphics[width=\wth\textwidth]{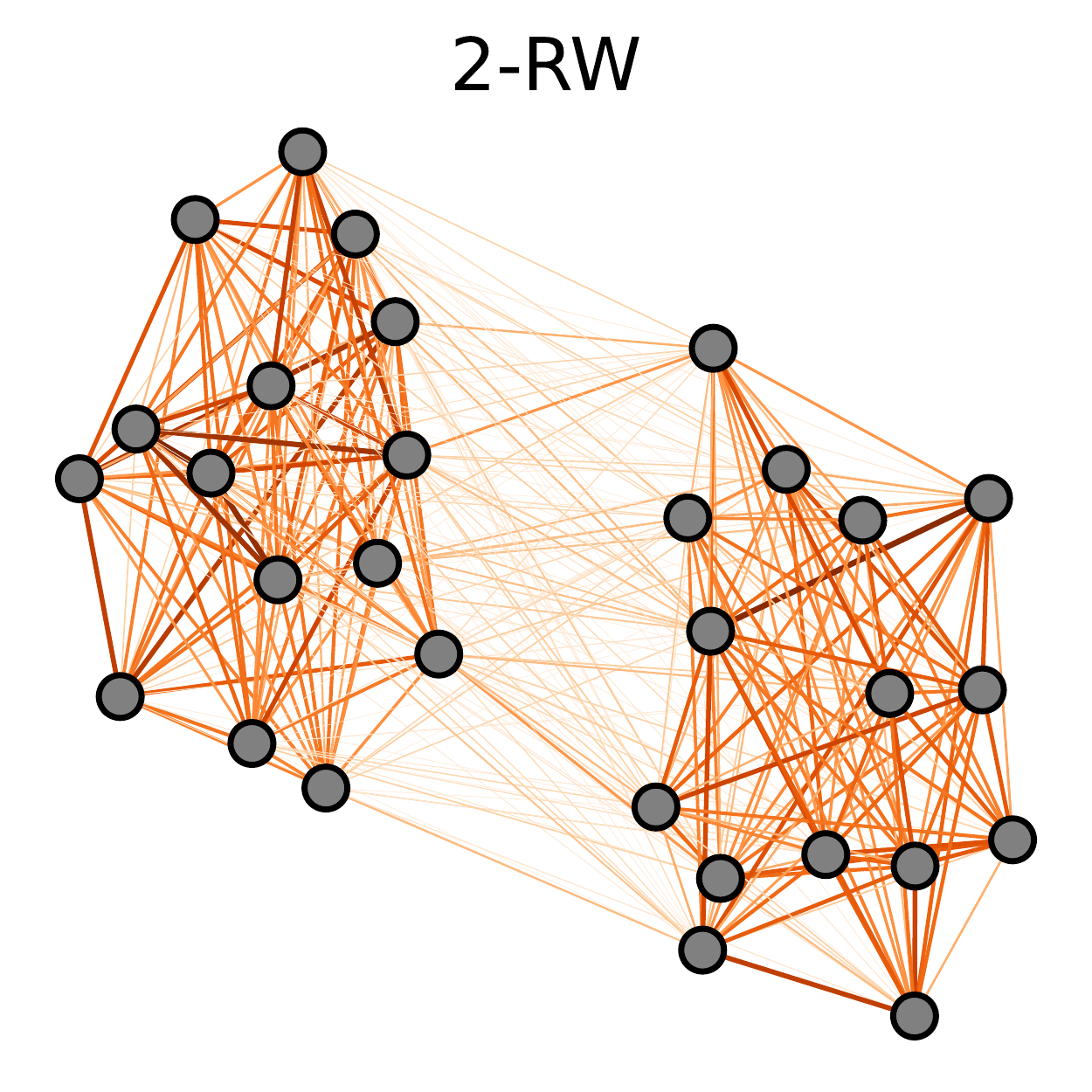}
\includegraphics[width=\wth\textwidth]{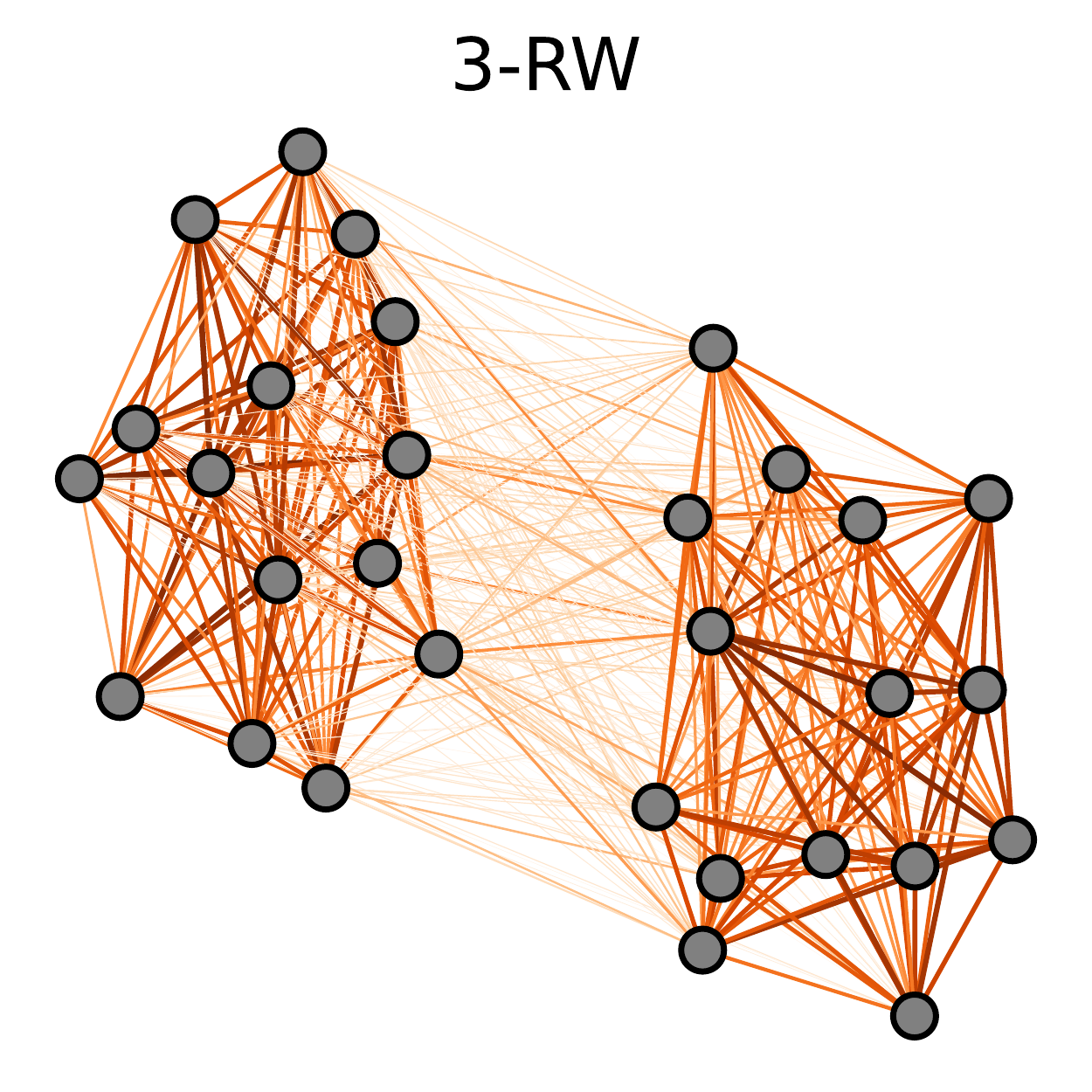}
\includegraphics[width=\wth\textwidth]{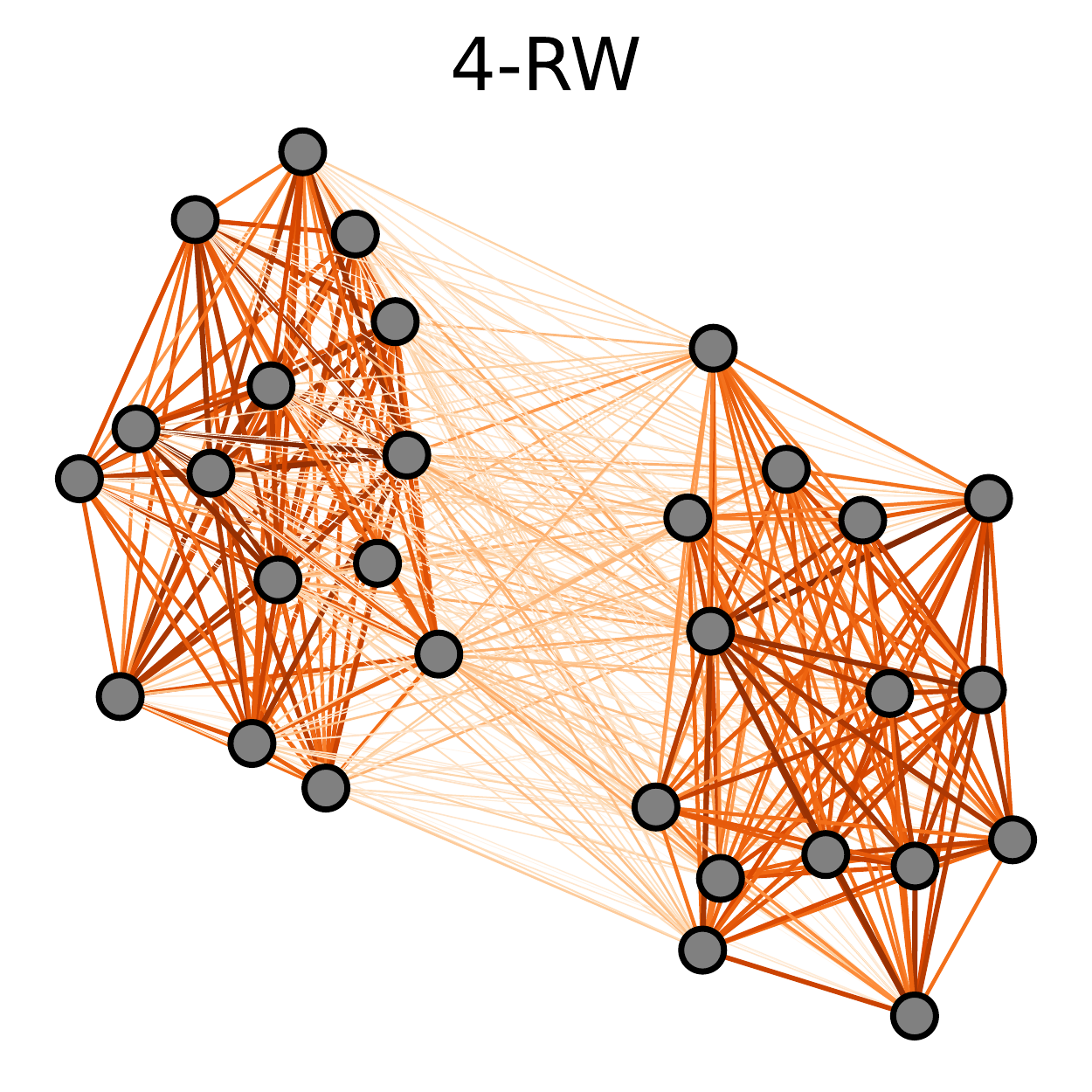}\\
}
\caption{RRWP visualization for a sample from a stochastic block model with 2 communities, up to the 4th power. Probabilities for longer random walks better highlight the community structure and reduce bottlenecks. }
\label{fig:rw_viz_sbm}
\end{figure*}

\section{Methodology and Theory}



In this section, we introduce our proposed GRIT architecture that uses a novel flexible attention mechanism together with a general relative positional encoding scheme, and does not incorporate any explicit local message-passing modules.

It is based on three design decisions: (i) a learned relative positional encoding initialized with random walk probabilities, (ii) a flexible attention mechanism that updates node and node-pair representations, 
and (iii) the integration of node degree information in each layer.
Each of our design decisions is justified by theoretical results, which are also covered in this section.
Further, in Section~\ref{sec:experiments}, we show that each of these design choices improves empirical performance in ablation studies, and the overall architecture achieves state-of-the-art performance across various datasets.

\subsection{Learned Random Walk Relative Encodings}


When applied to graph data, Transformers typically have a token embedding for each node and update this node embedding with attention and feedforward modules (FFNs). 
As the graph 
(adjacency) structure important to message-passing GNNs is removed from the architecture, the positional encodings must adequately capture the graph structure for Transformers to succeed.


We use a learned relative positional encoding scheme initialized with random walk probabilities that is related to previously used positional encodings~\citep{dwivedi2021GraphNeuralNetworks, li2020DistanceEncodingDesign, mialon2021GraphiTEncodingGraph}. Going beyond previous work, we give theoretical justification for the use of random walk probabilities --- with an appropriate architecture, they are more expressive than shortest path distances, and can capture large classes of graph propagation matrices.


\newcommand{\A}{\mathbf{A}}
\newcommand{\D}{\mathbf{D}}

Let $\A \in \RR^{n \times n}$ be the adjacency matrix of a graph $(\V, \E)$ with $n$ nodes, and let $\D$ be the diagonal degree matrix. 
Define $\rw := \D^{-1}\A$, and note that $\rw_{ij}$ is the probability that $i$ hops to $j$ in one step of a simple random walk.
The proposed relative random walk probabilities (RRWP) initial positional encoding is defined for each pair of nodes $i, j \in \V$ as follows:
\begin{equation}
    \P_{i,j} = [\mathbf{I}, \rw, \rw^2, \dots, \rw^{K-1}]_{i,j} \in \mathbb{R}^K,
\end{equation}

in which $\mathbf{I}$ is the identity matrix.
For any node $i \in \V$, the diagonal  $\P_{i,i}$ can additionally be utilized as an initial node-level structural encoding, which is the same as the Random Walk Structural Encodings (RWSE) used in past work~\citep{dwivedi2021GraphNeuralNetworks, rampasek2022RecipeGeneralPowerful}. 
The parameter $K \in \mathbb{N}$ controls the maximum length of random walks considered.

Importantly, we use RRWP as an initialization of learned relative positional encodings in our architecture. The tensor $\P$ can be updated by an elementwise $\mlp: \RR^K \to \RR^d$ to get new relative positional encodings $\mlp(\P_{i,j,:})$, and also is updated with other information by the attention layers in our Transformer. 
This is essential for expressive power, and also may provide a useful inductive bias, as in Proposition~\ref{prop:mlp_rrwp} we show that this updating allows us to recover important propagation matrices.







\paragraph{Visualization of RRWP}


To further justify and build intuition for initializing with RRWP positional encodings, we demonstrate that it indeed captures useful information on graphs by example.
In Figure~\ref{fig:rw_viz_mol} and Figure~\ref{fig:rw_viz_sbm} we visualize RRWP on two example graphs: the molecular graph for fluorescein and a sample graph from a stochastic block model with 2 communities. For the molecule, we see that RRWP captures higher-order structural information in the longer random walks. 
In the stochastic block model, longer random walks better reveal the community structure compared to the original graph topology. Moreover, it reduces bottlenecks in the graph, which may be important for certain applications~\citep{alon2020BottleneckGraphNeural, topping2022UnderstandingOversquashingBottlenecks}.

\subsubsection{Theory: RRWP + MLP is Expressive}

Given our initial RRWP positional encodings $\P$, we learn new positional encodings end-to-end with an $\mlp$.
We show that this combination is provably expressive: this learned positional encoding can approximate shortest path distances or general classes of graph propagation matrices up to an arbitrary $\epsilon > 0$ accuracy.
 This shows that we generalize methods like Graphormer~\cite{ying2021TransformersReallyPerform}, and various message-passing propagations~\citep{gasteiger2019DiffusionImprovesGraph, xu2019HowPowerfulAre}.
\begin{proposition}\label{prop:mlp_rrwp}
For any $n \in \mathbb{N}$, let $\mathbb{G}_n \subseteq \{0, 1\}^{n \times n}$ denote all adjacency matrices of $n$-node graphs. For $K \in \mathbb{N}$, and $\A \in \mathbb{G}_n$ consider the RRWP: $\P = [\mathbf{I}, \rw, \ldots, \rw^{K-1}] \in \RR^{n \times n \times K}$. Then for any $\epsilon > 0$, there exists an $\mlp: \RR^{K} \to \RR$ acting independently across each $n$ dimension such that $\mlp(\P)$ approximates any of the following to within $\epsilon$ error:
\begin{itemize}
    \item[(a)] $\mlp(\P)_{ij} \approx \spd_{K-1}(i,j)$ 
    \item[(b)] $\mlp(\P) \approx \sum_{k=0}^{K-1} \theta_k (\D^{-1} \A)^k$ 
    \item[(c)] $\mlp(\P) \approx \theta_0 \mathbf{I} + \theta_1 \A$,
\end{itemize}
in which $\spd_{K-1}(i,j)$ is the $K-1$ truncated shortest path distance, and $\theta_k \in \RR$ are arbitrary coefficients.
\end{proposition}
The proof is given in Appendix~\ref{appendix:rrwp_spd_prop}. 
(a) shows that, if we use an $\mlp$ with $K$-hop RRWP input, then we can capture all shortest path distances for nodes of up to $K-1$ hops away from each other. In particular, for $K = n$, we recover all shortest path distances (with disconnected nodes getting a distance of $n$, which is higher than the maximum distance $n-1$ between connected nodes). (b) and (c) show that we can capture many types of graph propagations. As special cases, we can capture sum aggregation [bullet (c) with  $\theta_0 = 0, \theta_1 = 1$], mean aggregation [bullet (b) with $K=2$, $\theta_1 = 1$, other $\theta_i = 0$], $K$-truncated personalized PageRank (PPR) [bullet (b) with $\theta_k = \alpha(1-\alpha)$ for some $\alpha \in (0,1)$], and $K$-truncated heat kernels [bullet (b) with $\theta_k = \exp(-\tau) \cdot \tau^k / k!$ for $\tau > 0$]. Such graph propagations may provide useful inductive biases, and a synthetic experiment in Section~\ref{sec:attn2mp} shows that GRIT is highly capable of learning attention mechanisms that match target graph propagations, while other Graph Transformers are less capable.


\subsection{Flexible Attention Mechanism with Absolute and Relative Representations}


Most current designs of the self-attention mechanism are based on node-token-level PE and node-token-level representations, but this does not fully capture the relative positional information between pairs of nodes.
A recent study by~\citet{brody2021HowAttentiveAre} also reveals that some attention mechanisms (e.g., GAT~\cite{velickovic2018GraphAttentionNetworks} and scaled dot-product attention~\cite{vaswani2017AttentionAllYou}) are not sufficiently \textit{flexible} to attend to specific tokens.

Therefore, we propose a new way to compute attention scores by conditioning on (learned) relative representations of node-pairs, which combines the strengths of the general conditioning layer~\cite{perez2018film} and GATv2~\cite{brody2021HowAttentiveAre}.

\newcommand{\x}{\mathbf{x}}
\newcommand{\e}{\mathbf{e}}
\newcommand{\W}{\mathbf{W}}

In each transformer layer, we update node representations $\x_i, \forall i \in \mathcal{V}$ and node-pair representations $\e_{i,j}, \forall i,j \in \mathcal{V}$. First, we initialize these using the initial node features and our RRWP positional encodings: $\x_i  = [\mathbf{x}'_i \| \P_{i,i}] \in \RR^{d_h + K}$ and $\e_{i,j} = [\e'_{i,j} \| \P_{i,j}] \in \RR^{d_e + K}$, where $\x'_i \in \RR^{d_h}$ and $\e'_{i,j} \in \RR^{d_e}$ are observed node and edge attributes, which can be dropped if not present in the data.
We set $\e'_{i,j} = \mathbf{0}$ if there is no observed edge from $i$ to $j$ in the original graph.
The attention computation is defined as follows:
\begin{equation}
   \begin{aligned}
    &
    \hat{\e}_{i,j} = \sigma\Big( \rho\left(\left(\W_\text{Q} \x_i + \W_\text{K}\x_j\right) \odot \W_\text{Ew}\e_{i,j}\right)\\
   & \text{\hspace{5cm}} + \W_\text{Eb}\e_{i,j} \Big) \in \mathbb{R}^{d'}, \\
    &\alpha_{ij} = \text{Softmax}_{j \in \mathcal{V}} (\W_\text{A} \hat{\e}_{i,j}) \in \mathbb{R},\\
    & \hat{\x}_i =  \sum_{j \in \mathcal{V}} \alpha_{ij} \cdot ( \W_\text{V} \x_j + \W_\text{Ev} \hat{\e}_{i,j}) \in \mathbb{R}^d,\\
   \end{aligned}
\end{equation}
where $\sigma$ is a non-linear activation (ReLU by default); 
$\W_\text{Q}, \W_\text{K}, \W_\text{Ew}, \W_\text{Eb} \in \mathbb{R}^{d' \times d}$, $\W_\text{A} \in \mathbb{R}^{1 \times d'}$ and $\W_\text{V}, \W_\text{Ev} \in \mathbb{R}^{d \times d'}$ are learnable weight matrices; $\odot$ indicates elementwise multiplication; and $\rho(\x):= (\text{ReLU}(\x))^{1/2}-(\text{ReLU}(-\x))^{1/2}$ is the signed-square-root, which stabilizes training by reducing the magnitude of large inputs.
We also include biases in our implementation, but they are omitted here for simplicity.
Note that we update the pair representations $\e_{i,j}$, so our Transformer is capable of updating the positional encodings. In particular, our Transformer is capable of applying an elementwise $\mlp$ to $\P$, as we showed was useful in Proposition~\ref{prop:mlp_rrwp}.

Similarly to other self-attention mechanisms, our proposed attention mechanism can be extended to multiple heads (say, $N_h$ heads) by assigning different weight matrices for different heads. We perform the above computations for different heads $h \in \{1, \ldots, N_h\}$ to get representations $\hat{\x}^h_i$ and $\hat{\e}^h_{i,j}$, then combine the different heads as follows:
\begin{equation}\label{eq:attention_out}
   \begin{aligned}
    \x^\text{out}_i =  \sum_{h=1}^{N_h} \W^h_\text{O}\hat{\x}^h_i \in \mathbb{R}^d\,,  \\
    \e^\text{out}_{ij} =  \sum_{h=1}^{N_h} \W^h_\text{Eo}\hat{\e}^h_{ij}  \in \mathbb{R}^d \,, \\
   \end{aligned}
\end{equation}
where $\W^h_\text{O}, \W^h_\text{Eo}  \in \mathbb{R}^{d \times d'}$ are output weight matrices for each head $h$.

\newcommand{\hash}{\mathrm{hash}}

\subsubsection{Theory: RRWP is more expressive than SPD in Transformers}

\begin{figure}[t!]
    \centering
    \includegraphics[width=.4\columnwidth]{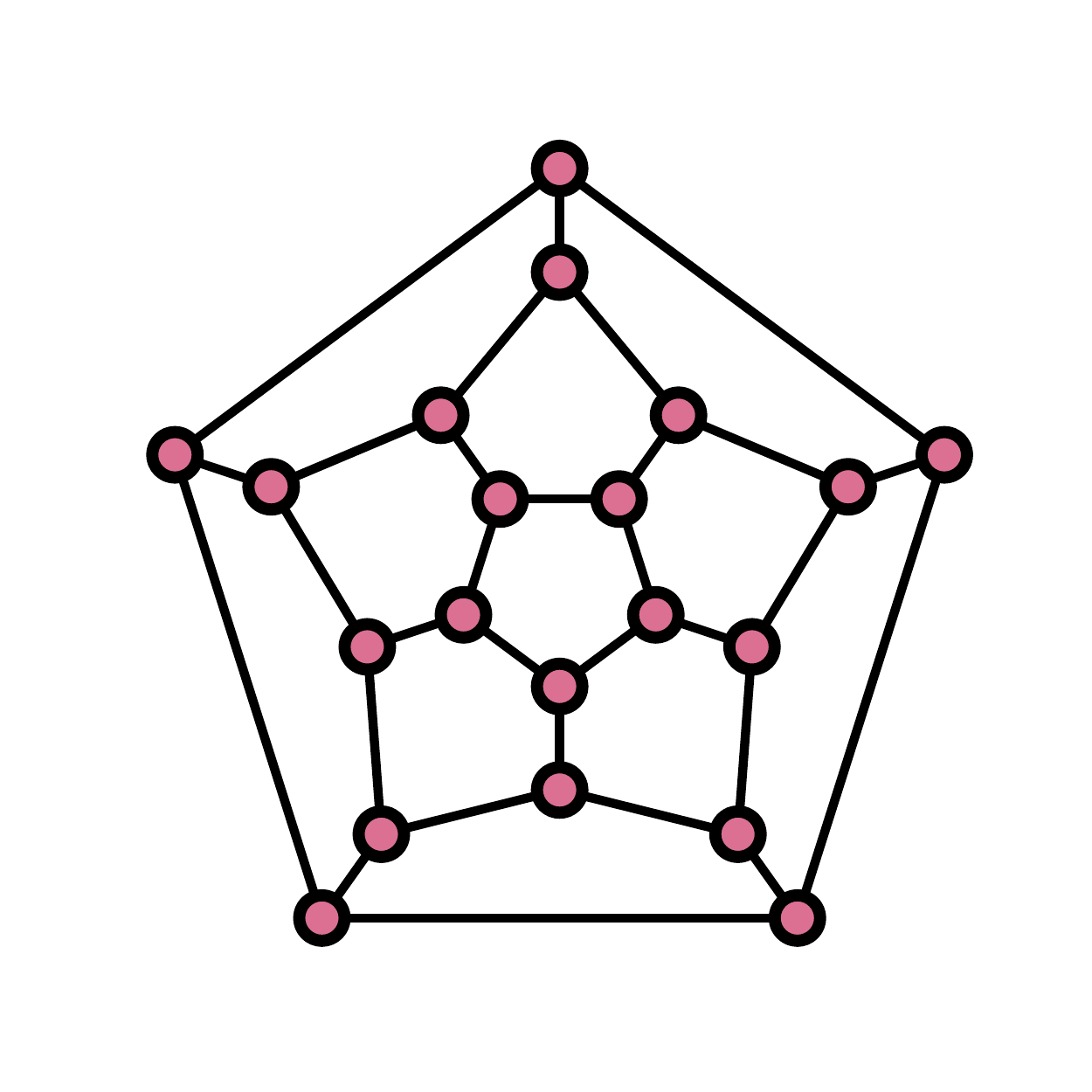}
    \includegraphics[width=.4\columnwidth]{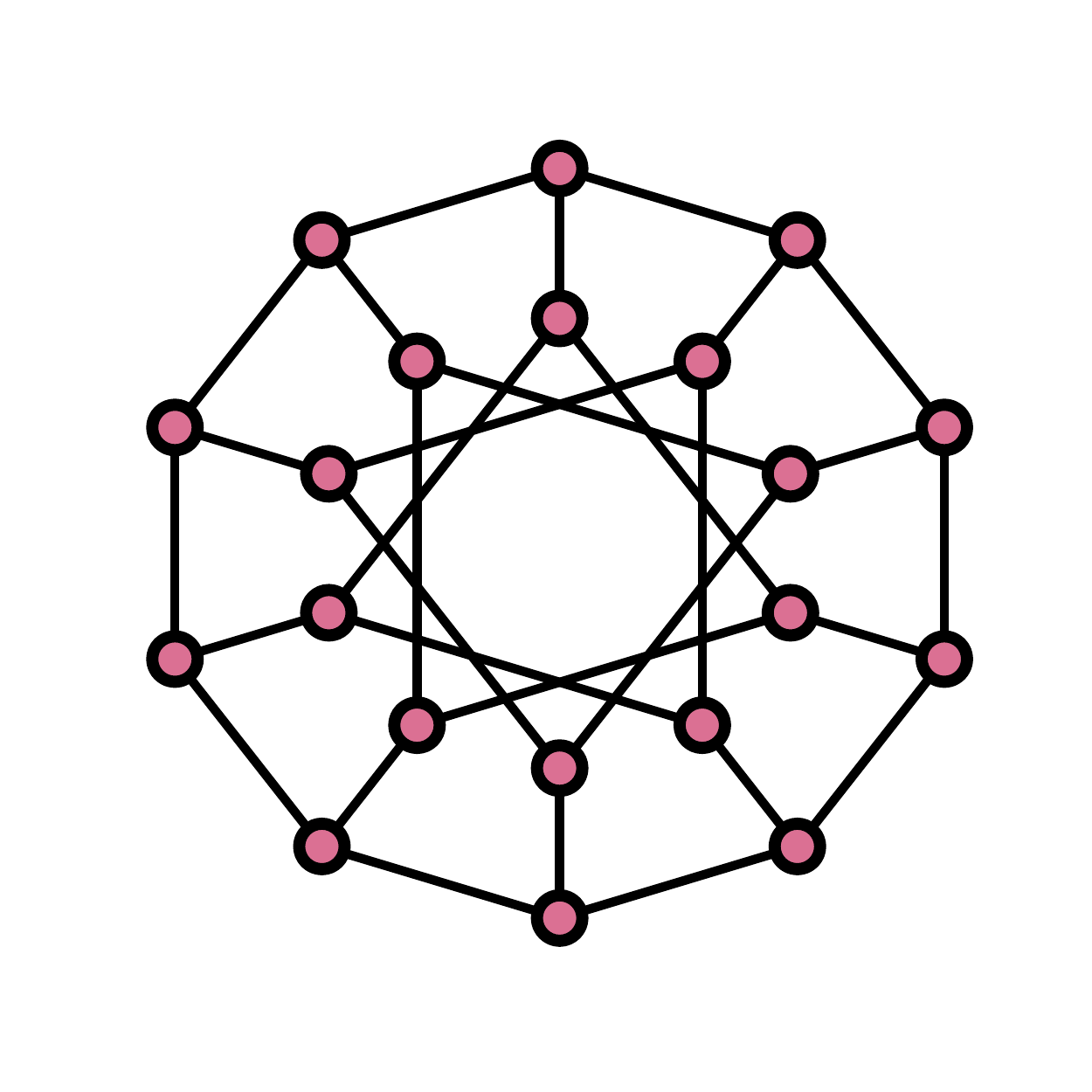}
    \caption{(Left) Dodecahedron graph. (Right) Desargues graph. GD-WL with RRWP can distinguish these two graphs, but GD-WL with SPD cannot.}
    \label{fig:dd_graphs}
\end{figure}

We can use recently proposed Weisfeiler-Leman-like graph isomorphism tests to demonstrate that RRWP within a Transformer architecture is strictly more expressive than the commonly used shortest path distances (SPD).
Recently, \citet{zhang2023rethinking} proposed the Generalized Distance Weisfeiler-Leman Test (GD-WL) --- a graph isomorphism test based on updating node colors that incorporates distances. Let $G$ be a graph with vertex set $\V$, $d_G(v, u)$ denote a distance between nodes $v$ and $u$, and $\chi_G^0(v)$ be an initial color of $v$. Then GD-WL updates node colors as
\newcommand{\multiset}[1]{\{\!\!\{#1\}\!\!\}}
\begin{equation}
    \chi_G^t(v) = \hash(\multiset{ (d_G(v,u),  \chi_G^{t-1}(u)) : u \in \V}  ).
\end{equation}
The multiset of final node colors $\multiset{\chi_G^T(v): v \in \V}$ at iteration $T$ is hashed to get a graph color. \citet{zhang2023rethinking} use GD-WL to analyze a Graph Transformer architecture that uses $d_G(v,u)$ as relative positional encodings. They show that setting $d_G(v,u)$ to the shortest path distance makes it possible to solve edge biconnectivity problems. We can show that if we choose $d_G(v,u)$ to be our relative random walk encoding $d_G(v,u) = \P_{vu} \in \RR^n$ (with $K=n$), the GD-WL using this distance is more powerful than the GD-WL using shortest path distances (SPD)
\footnote{We need to generalize GD-WL to allow vector-valued distances, but this is easily handled by the $\hash$ function}.
\begin{proposition}\label{prop:rrwp_gdwl}
    GD-WL with RRWP distances is strictly stronger than GD-WL with shortest path distances.
\end{proposition}
The proof is given in Appendix~\ref{appendix:gdwl_proof}. We first show that GD-WL with RRWP can distinguish any two graphs that GD-WL with SPD can. Then we show that GD-WL with RRWP can distinguish the Dodecahedron and Desargues graphs --- plotted in Figure~\ref{fig:dd_graphs} --- whereas \citet{zhang2023rethinking} showed that GD-WL with SPD cannot distinguish these.

\subsection{Injecting Degree Information}

Attention mechanisms are innately invariant to node degrees, analogously to mean-aggregation; this introduces extra ambiguities and hence reduces expressive power in processing graph-structured data~\cite{xu2019HowPowerfulAre, corso2020PrincipalNeighbourhoodAggregation}.
Therefore, we introduce an adaptive degree-scaler~\citep{corso2020PrincipalNeighbourhoodAggregation} to our attention mechanism to maintain degree information.

After the computation of node representations in \eqref{eq:attention_out}, we inject degree information into the node representations as follows:
\newcommand{\btheta}{\boldsymbol{\theta}}
\begin{equation}
   \begin{aligned}
    \x^{\text{out}'}_i :=  \x^\text{out}_i \odot \btheta_1 + \left( \log(1 + d_i) \cdot \x^\text{out}_i \odot \btheta_2 \right) \in \mathbb{R}^d \,, \\
   \end{aligned}
\end{equation}
where $d_i$ is the degree of node $i$, and $\btheta_1, \btheta_2 \in \mathbb{R}^d$ are learnable weights. Like other Transformer architectures, we follow this with a standard feed-forward network (FFN) to update the node representations. 

To properly include degree information, we apply batch normalization~\cite{ioffe2015BatchNormalizationAccelerating} instead of the standard layer normalization~\cite{ba2016LayerNormalization, vaswani2017AttentionAllYou} to the outputs of self-attention modules and FFNs.  
This is because layer normalization applied to each node representation cancels out the effect from degree-scalers or sum-aggregators. We capture this in the following proposition, which we prove in Appendix~\ref{appendix:layernorm_degree}.
\begin{proposition}\label{prop:ln_degree}
    Sum-aggregated node representations, degree-scaled node representations, and mean-aggregated node representations all have the same value after application of a LayerNorm on node representations.
\end{proposition}



\section{Experimental Results}\label{sec:experiments}

\newcommand{\first}[1]{\textcolor{SeaGreen}{#1}}
\newcommand{\second}[1]{\textcolor{BurntOrange}{#1}}
\newcommand{\third}[1]{\textcolor{Periwinkle}{#1}}

\begin{table*}[h!]
    \centering
    \caption{
    Test performance in five benchmarks from \cite{dwivedi2020BenchmarkingGraphNeural}. 
    Shown is the mean $\pm$ s.d. of 4 runs with different random seeds. Highlighted are the top \first{first}, \second{second}, and \third{third} results. 
    \# Param $\sim500K$ for ZINC, PATTERN, CLUSTER and $\sim 100K$ for MNIST and CIFAR10.
    ${}^*$ indicates statistically significant difference against the second-best result from the two-sample one-tailed t-test.}
    {\small
    \begin{tabular}{lccccc}
    \toprule
       \textbf{Model}  &  \textbf{ZINC} & \textbf{MNIST} & \textbf{CIFAR10} & \textbf{PATTERN} & \textbf{ CLUSTER} \\
       \cmidrule{2-6} 
       & \textbf{MAE}$\downarrow$  &  \textbf{Accuracy}$\uparrow$ & \textbf{Accuracy}$\uparrow$ & \textbf{Accuracy}$\uparrow$ & \textbf{Accuracy}$\uparrow$ \\
       \midrule
       GCN  & $0.367 \pm 0.011$ & $90.705 \pm 0.218$ & $55.710 \pm 0.381$ & $71.892 \pm 0.334$ & $68.498 \pm 0.976$ \\
GIN  & $0.526 \pm 0.051$ & $96.485 \pm 0.252$ & $55.255 \pm 1.527$ & $85.387 \pm 0.136$ & $64.716 \pm 1.553$ \\
GAT & $0.384 \pm 0.007$ & $95.535 \pm 0.205$ & $64.223 \pm 0.455$ & $78.271 \pm 0.186$ & $70.587 \pm 0.447$ \\
GatedGCN & $0.282 \pm 0.015$ & $97.340 \pm 0.143$ & $67.312 \pm 0.311$ & $85.568 \pm 0.088$ & $73.840 \pm 0.326$ \\
GatedGCN-LSPE & $0.090 \pm 0.001$ & $-$ & $-$ & $-$ & $-$ \\
PNA & $0.188 \pm 0.004$ & $97.94 \pm 0.12$ & $70.35 \pm 0.63$ & $-$ & $-$ \\
DGN & $0.168 \pm 0.003$ & $-$ &  \second{$\mathbf{72.838 \pm 0.417}$} & $86.680 \pm 0.034$ & $-$ \\
GSN  & $0.101 \pm 0.010$ & $-$ & $-$ & $-$ & $-$ \\
\midrule CIN & \third{$\mathbf{0.079} \pm \mathbf{0.006}$} & $-$ & $-$ & $-$ & $-$ \\
CRaW1 & $0.085 \pm 0.004$ & ${97.944} \pm {0.050}$ & $69.013 \pm 0.259$ & $-$ & $-$ \\
GIN-AK+ & ${0 . 0 8 0} \pm {0 . 0 0 1}$ & $-$ & $72.19 \pm 0.13$ & \second{$\mathbf{86.850 \pm 0.057}$} & $-$ \\
\midrule SAN & $0.139 \pm 0.006$ & $-$ & $-$ & $86.581 \pm 0.037$ & $76.691 \pm 0.65$ \\
Graphormer & $0.122 \pm 0.006$ & $-$ & $-$ & $-$ & $-$ \\
K-Subgraph SAT & $0.094 \pm 0.008$ & $-$ & $-$ & \third{$\mathbf{86.848 \pm 0.037}$} & $77.856 \pm 0.104$ \\
EGT & $0.108 \pm 0.009$ & \first{$\mathbf{98.173 \pm 0.087}$} & $68.702 \pm 0.409$ & $86.821 \pm 0.020$ &  \second{$\mathbf{79.232 \pm 0.348}$} \\
Graphormer-URPE & $0.086 \pm 0.007$ & $-$ & $-$ & $-$ & $-$  \\
Graphormer-GD & $0.081 \pm 0.009$ & $-$ & $-$ & $-$ & $-$ \\
 GPS & \second{$\mathbf{0.070} \pm \mathbf{0.004}$} & \third{$\mathbf{98.051 \pm 0.126}$} & \third{$\mathbf{72.298 \pm 0.356}$} & $86.685 \pm 0.059$ & \third{$\mathbf{78.016 \pm 0.180}$} \\
\midrule
\ourmethod (ours) & \first{$\mathbf{0.059 \pm 0.002}^* $}&  \second{$\mathbf{98.108  \pm 0.111}$} & \first{$\mathbf{76.468 \pm 0.881}^*$} & \first{$\mathbf{87.196 \pm 0.076}^*$}  & \first{$\mathbf{80.026 \pm 0.277}^*$} \\
       \bottomrule
    \end{tabular}
    }\\
    \label{tab:exp_bmgnn}
\end{table*}

\begin{table}[h!]
    \caption{Test performance on two benchmarks from long-range graph benchmarks (LRGB)~\cite{dwivedi2022LongRangeGraph}. 
    Shown is the mean $\pm$ s.d. of 4 runs with different random seeds. Highlighted are the top \first{first}, \second{second}, and \third{third} results.
    \#~Param $\sim 500K$ for both datasets.
     ${}^*$ indicates statistical significance against the second-best results from the Two-sample One-tailed T-Test. 
    }
    \setlength{\tabcolsep}{2pt}
    {\small
    \begin{tabular}{lcc}
    \toprule
       \textbf{Model}  &  \textbf{Peptides-func} & \textbf{Peptides-struct} \\
       \cmidrule{2-3} 
       &  \textbf{AP}$\uparrow$ & \textbf{MAE}$\downarrow$ \\
       \midrule
       GCN  & $0.5930 \pm 0.0023$  & $0.3496 \pm 0.0013$ \\
GINE & $0.5498 \pm 0.0079$ & $0.3547 \pm 0.0045$ \\
GatedGCN & $0.5864 \pm 0.0035$ & $0.3420 \pm 0.0013$ \\
GatedGCN+RWSE & $0.6069 \pm 0.0035$ & $0.3357 \pm 0.0006$ \\
\midrule 
Transformer+LapPE & $0.6326 \pm 0.0126$ & \third{$\mathbf{0.2529 \pm 0.0016}$} \\
SAN+LapPE & $0.6384 \pm 0.0121$  & $0.2683 \pm 0.0043$ \\
SAN+RWSE& \third{$\mathbf{0.6439\pm 0.0075}$}  & $0.2545\pm 0.0012$ \\
GPS & \second{$\mathbf{0.6535 \pm 0.0041}$} & \second{$\mathbf{0.2500 \pm 0.0012}$}\\
\midrule
\ourmethod (ours) & \first{$\mathbf{0.6988 \pm 0.0082}^*$} & \first{$\mathbf{0.2460 \pm 0.0012}^*$}
\\
       \bottomrule
    \end{tabular}
    }
    \label{tab:lrgb}
\end{table}

\subsection{Benchmarking \ourmethod}

We evaluate our proposed method on five benchmarks from the Benchmarking GNNs work~\cite{dwivedi2020BenchmarkingGraphNeural} and two benchmarks from the recently developed Long-Range Graph Benchmark~\cite{dwivedi2022LongRangeGraph}. 
These benchmarks cover diverse graph learning tasks including node classification, graph classification, and graph regression; they are especially focused on graph structure encoding, node clustering, and learning long-range dependencies.  In addition, we also conduct experiments on the larger datasets ZINC-full graphs ($\sim$ 250,000 graphs)~\cite{irwin2012ZINCFreeTool} and PCQM4Mv2 ($\sim$ 3,700,000 graphs)~\citep{hu2021ogblsc}.

Further details concerning the experimental setup can be found in Appendix~\ref{appendix:experiment_details}.



\paragraph{Baselines}
We primarily compare our methods with the recent SOTA hybrid Graph Transformer, GraphGPS~\cite{rampasek2022RecipeGeneralPowerful},
as well as a number of prevalent graph-learning models:
popular message-passing neural networks (GCN~\cite{kipf2017SemiSupervisedClassificationGraph}, GIN~~\cite{xu2019HowPowerfulAre} and its variant with edge-features~\cite{hu2020StrategiesPretrainingGraph}, GAT~\cite{velickovic2018GraphAttentionNetworks}, GatedGCN~\cite{bresson2018ResidualGatedGraph}, GatedGCN-LSPE~\cite{dwivedi2021GraphNeuralNetworks},
PNA~\cite{corso2020PrincipalNeighbourhoodAggregation}); 
Graph Transformers (Graphormer~\cite{ying2021TransformersReallyPerform}, 
K-Subgraph SAT~\cite{chen2022StructureAwareTransformerGraph}, EGT~\cite{hussain2022GlobalSelfAttentionReplacement}, SAN~\cite{kreuzer2021RethinkingGraphTransformers}, 
Graphormer-URPE~\cite{luo2022your},
Graphormer-GD~\cite{zhang2023rethinking}); and other recent Graph Neural Networks with SOTA performance (DGN~\cite{beani2021DirectionalGraphNetworks},
GSN~\cite{bouritsas2022ImprovingGraphNeural}, 
CIN~\cite{bodnar2021WeisfeilerLehmanGo},
CRaW1~\cite{toenshoff2021graph},
GIN-AK+~\cite{zhao2021stars}).

\paragraph{Benchmarks from Benchmarking GNNs~\cite{dwivedi2020BenchmarkingGraphNeural}.} 
We first benchmark our method on five datasets from Benchmarking GNNs~\cite{dwivedi2020BenchmarkingGraphNeural}: ZINC, MNIST, CIFAR10, PATTERN, and CLUSTER, following the experimental setting of GraphGPS~\cite{rampasek2022RecipeGeneralPowerful} ($\sim500$K parameter limit for ZINC, PATTERN, and CLUSTER; $\sim100$K parameter limit for MNIST and CIFAR10). Results are shown in Table~\ref{tab:exp_bmgnn}; we report the mean and standard deviation across 4 runs with different random seeds.


We show that our model has the best mean performance for four of the five datasets with statistically significant improvement.
In the remaining dataset, our model reaches the second-best performance without a statistically significant difference compared to the best performer.
These results showcase the ability of GRIT to outperform a variety of methods on small to medium-sized datasets, including MPNNs, expressive higher-order GNNs, and Graph Transformers.

 \paragraph{Long-Range Graph Benchmark~\cite{dwivedi2022LongRangeGraph}.}
 Next, we evaluate our method on the recently proposed Long-Range Graph Benchmark (LRGB).
 We conduct experiments on the two peptide graph benchmarks from LRGB, namely Peptides-func and Peptides-struct, which are 10-task multilabel classification and 11-task regression tasks, respectively.
 Results are shown in Table~\ref{tab:lrgb}.
On both datasets, our method obtains the best mean performance, outperforming MPNNs and Graph Transformers --- this demonstrates that our model is capable of learning long range interactions.


\paragraph{ZINC-full Dataset}

We also test our model on the ZINC-full dataset~\cite{irwin2012ZINCFreeTool}, which is the full version of ZINC that has 250,000 graphs.
Besides the MPNNs and Graph Transformers, we also compare our method with other domain agnostic methods like higher-order GNNs ($\delta$-2-GNN, $\delta$-2-LGNN~\cite{morris2020weisfeiler}) as well as PE-enhanced GNNs (SignNet~\cite{lim2022SignBasisInvariant}).
We see that \ourmethod outperforms various classes of methods, and achieves the best mean performance of all methods.

\begin{table}[h!]
    \caption{Test performance on ZINC-full~\cite{irwin2012ZINCFreeTool}.
    \#~Param $\sim500K$.
    Shown is the mean $\pm$ s.d. of 4 runs with different random seeds. Highlighted are the top \first{first}, \second{second}, and \third{third} results.
    }
    {\small
    \begin{tabular}{clc}
    \toprule
       \textbf{Method} & \textbf{Model}  & 
       \textbf{ZINC-full} (MAE $\downarrow$) \\ \midrule
        \multirow{5}{*}{MPNNs}
       
       & GIN & $0.088 \pm 0.002$ \\
       & GraphSAGE & $0.126 \pm 0.003$ \\
       & GAT & $0.111 \pm 0.002$ \\
       & GCN & $0.113 \pm 0.002$ \\
       & MoNet & $0.090 \pm 0.002$ \\
       \midrule
       Higher-order & $\delta$-2-GNN & $0.042 \pm 0.003 $ \\
       GNNs & $\delta$-2-LGNN & $0.045 \pm 0.006$ \\
       \midrule
       PE-GNN & SignNet & $\second{\mathbf{0.024 \pm 0.003}}$ \\
       \midrule 
       & Graphormer & $0.052 \pm 0.005$ \\
       Graph & Graphormer-URPE  & $0.028 \pm 0.002$ \\
       Transformers & Graphormer-GD & $\third{\mathbf{0.025 \pm 0.004}}$ \\
       \cmidrule{2-3} 
       & \ourmethod (ours) & $\first{\mathbf{0.023 \pm 0.001}}$  \\
       \bottomrule
    \end{tabular}
    }
    \label{tab:zinc_full}
\end{table}

\paragraph{PCQM4Mv2 Large-scale Graph Regression Benchmark~\cite{hu2021ogblsc}}

Further, we conduct an experiment on the PCQM4Mv2 large-scale graph regression benchmark of 3.7M~graphs~\cite{hu2021ogblsc},
which is currently one of the largest molecular datasets (Table.~\ref{tab:pcqm4mv2}).
We compare our method against MPNNs (GCN~\cite{kipf2017SemiSupervisedClassificationGraph}, GIN~\cite{xu2019HowPowerfulAre} with/without virtual nodes) as well as several Graph Transformers (GRPE~\cite{park2022GRPERelativePositional}, Graphormer~\cite{ying2021TransformersReallyPerform}, TokenGT~\cite{kim2022PureTransformersAre} and GraphGPS~\cite{rampasek2022RecipeGeneralPowerful}).
Following the protocol of \citet{rampasek2022RecipeGeneralPowerful},  we treated the validation set of
the dataset as a test set, since the \textit{Test-dev} set labels are private.
The result of a single random seed run is reported due to the size of the dataset, following previous works~\cite{rampasek2022RecipeGeneralPowerful, kim2022PureTransformersAre}.
Our model can reach a comparable performance to GraphGPS (best) and Graphormer (third best) while using fewer learnable parameters. 
We did not conduct any hyperparameter search due to the limit of time, 
and instead adopted the values from GraphGPS~\cite{rampasek2022RecipeGeneralPowerful}.

\begin{table}[h!]
    \centering
    \caption{Test performance on PCQM4Mv2~\cite{hu2021ogblsc} dataset.
    Shown is the result of a single run due to the computation constraint. Highlighted are the top \first{first}, \second{second}, and \third{third} results.
    }
    \setlength{\tabcolsep}{2pt}
    {\small
    \begin{tabular}{clcc}
    \toprule
       \textbf{Method} & \textbf{Model}  & 
       \textbf{Valid.} (MAE $\downarrow$) & \textbf{\#~Param} \\ \midrule
        \multirow{5}{*}{MPNNs}
       & GCN & $0.1379$ & 2.0M\\
       & GCN-virtual & $0.1153$ & 4.9M\\
       & GIN & $0.1195$ & 3.8M \\ 
       & GIN-virtual & $0.1083$ & 6.7M\\ 
       \midrule
       & GRPE & $0.0890$ & 46.2M \\
        & Graphormer & $\third{\mathbf{0.0864}}$ & 48.3M \\
       & TokenGT (ORF) & 0.0962 & 48.6M \\
      Graph      & TokenGT (Lap) & 0.0910 & 48.5M \\
       Transformers & GPS-small & $0.0938$ & 6.2M \\
      & GPS-medium & $\first{\mathbf{0.0858}}$ & 19.4M\\
       \cmidrule{2-4} 
       &  \ourmethod (ours)& $\second{\mathbf{0.0859}}$ & 16.6M \\
       \bottomrule
    \end{tabular}
    }
    \label{tab:pcqm4mv2}
\end{table}


\begin{table}[h!]
    \centering
    \caption{Ablations on design choices of our architecture on ZINC~\cite{dwivedi2020BenchmarkingGraphNeural}. Substituting other design choices decreases the performance of our model.
    Shown is the mean $\pm$ s.d. of 4 runs with different random seeds. A $\to$ B stands for using B instead of A.
    }
\begin{tabular}{lc}
\toprule
   ZINC & MAE $\downarrow$    \\
   \midrule
    \ourmethod (ours) &  $\mathbf{0.059 \pm 0.002}$ \\ 
    \ - Remove degree scaler & $0.076 \pm 0.002$   \\
    \ - Remove the update of RRWP & $0.066 \pm 0.005$ \\
    \ - Global-attn. $\to$ Sparse-attn. & $0.066 \pm 0.002$ \\
    \ - Degree scaler $\to$ Degree encoding  & $0.072 \pm 0.005$    \\
    \ - GRIT-attn. $\to$ Graphormer-attn. & $0.117 \pm 0.028$  \\
    \ - RRWP $\to$ RWSE  &  $0.081 \pm 0.010$ \\
    \ - RRWP $\to$ SPDPE & $0.067 \pm 0.002$ \\
    \bottomrule
\end{tabular}
    \label{tab:ablation}
\end{table}

\subsection{Ablations}\label{sec:ablations}

To determine the utility of our architectural design choices, we conduct several ablation experiments on ZINC. Table~\ref{tab:ablation} shows the results. 
Removing degree scalers, 
removing the update mechanism of RRWP, 
substituting global attention with sparse attention,
replacing the degree scalers with degree encoding from Graphormer~\cite{ying2021TransformersReallyPerform},
replacing our attention mechanism with the attention mechanism used in Graphormer (no PE update),
and substituting RRWP with RWSE~\cite{dwivedi2021GraphNeuralNetworks} or SPDPE~\cite{ying2021TransformersReallyPerform}, all lead to worse performance --- this lends credence to our architectural choices.

We also conduct a sensitivity analysis on the parameter $K$ of RRWP on ZINC. The results are shown in Table~\ref{tab:rrwp_k}.
Notably, our method is SOTA or near SOTA for many choices of $K$, except for very unreasonable choices like $K=2$.
Note that we keep every other hyperparameter fixed besides $K$ in this experiment, which was originally chosen for $K=21$. This might explain why other $K$ perform slightly worse.

\begin{table}[ht]
    \centering
    \setlength{\tabcolsep}{2pt}
    \caption{Sensitivity Analysis of $K$-order RRWP on ZINC~\cite{dwivedi2020BenchmarkingGraphNeural}. Shown is the mean $\pm$ s.d. of 4 runs with different random seeds.}
    \label{tab:rrwp_k}
    {\small
\begin{tabular}{l|ccccccc}
    \toprule
    $K$ &  $2$ & $7$ & $14$ & $18$ & $21$ & $24$ & $42$ \\ \midrule
    MAE $\downarrow$ & $0.147$ & $0.063$ & $0.063$ & $0.060$ & $0.059$ & $0.060$ & $0.061$ \\
    \quad \quad $\pm$  & $0.006$ & $0.003$ & $0.004$ & $0.002$ & $0.002$ & $0.002$ & $0.001$ \\
    \bottomrule
    \end{tabular}
    }
\end{table}


\begin{table}[ht]
    \centering
    \setlength{\tabcolsep}{2pt}
    \caption{Synthetic Experiment on Learning to Attend $K$-hop Neighborhoods}
    \label{tab:attn2mp}
    {\small
\begin{tabular}{l|lll}
    \toprule
        MAE $\leq 1$ $\downarrow$ & 1-hop & 2-hop & 3-hop  \\ \midrule
        MeanPool (baseline) & $ .083 \pm .015 $ & $.080 \pm .014$ & $.069 \pm .011$  \\ \midrule
        Transf.+RWSE & $.083 \pm .015$ & $.080 \pm .014$ & $.069 \pm .011$  \\ 
        SAN+LapPE & $.044 \pm .011$ & $.042 \pm .010$ & $.029 \pm .008$  \\ 
        Graphormer+SPDPE & $ .043 \pm .010$ & $.034 \pm .010$ & $.025 \pm .005$  \\ \midrule
        GRIT (Ours) & $\first{\mathbf{.001 \pm .001}}$ & $\first{\mathbf{.001 \pm .001}}$ & $\first{\mathbf{.007 \pm .004}}$ \\ \bottomrule
    \end{tabular}
    }
\end{table}


\subsection{Synthetic Experiment: Can Our Attention Module Learn to Attend to $K$-hop Neighbors?}
\label{sec:attn2mp}


We conduct a synthetic experiment to study the ability 
of our proposed attention modules 
to emulate a general class of graph propagation matrices, in comparison to existing Graph Transformers.
Concretely, we consider single-layer, single-head attention modules with the corresponding positional encoding as the only input.
We experiment with GRIT and a set of popular Graph Transformers:
vanilla Transformer+RWSE (the Transformer branch of GraphGPS~\cite{rampasek2022RecipeGeneralPowerful}), SAN+LapPE*~(global attention component with the LapPE transformer)~\cite{kreuzer2021RethinkingGraphTransformers}, 
and Graphormer+SPDPE~\cite{ying2021TransformersReallyPerform}.

We train these attention mechanisms to output a row-normalized adjacency matrix for $(k=1, 2, 3)$-hop neighborhoods, using the $l_1$-loss as the training objective. More precisely, the targets are generated as follows. We choose a $k \in \{1, 2, 3\}$, compute $\A^k$ for an adjacency matrix $\A$, round nonzero values to have the value $1$, and then normalize each row to sum to one.

With $20$ graphs randomly sampled from the ZINC dataset, we train each method for 2000 epochs on each graph separately and compute the (training) mean absolute errors across graphs.
From the results (Table~\ref{tab:attn2mp}), we observe that SAN+LapPE* and Graphormer+SPDPE perform similarly in learning to attend to a specific type of neighborhood,
while Transformer+RWSE performs much worse; this may help explain why MPNNs are so essential for GraphGPS, according to its ablation studies (Table B.1~\cite{rampasek2022RecipeGeneralPowerful}).
GRIT significantly outperforms other baselines by an order of magnitude, 
showing that our method can well approximate a general class of graph propagations, as theoretically expected (due to Proposition~\ref{prop:mlp_rrwp}).





In addition, we also visualize the attention scores for a single graph of the synthetic experiments (see Figure~\ref{fig:visual_attn} in Appendix~\ref{appendix:synthetic}), which agrees with the quantitative results. 
Whereas other attention mechanisms qualitatively struggle at matching the target sparsity pattern or attention magnitudes, our GRIT attention mechanism succeeds at learning to match both sparsity and magnitudes.

\section{Conclusion}
Observing the performance gap of Graph Transformers between small and large-scale datasets, 
we argue for the importance of graph inductive biases in Graph Transformers.
Motivated by promoting inductive biases in Graph Transformers without message-passing, 
we propose GRIT, based on three theoretically and empirically motivated design choices for incorporating graph inductive biases.
Our learned relative positional encodings initialized with RRWP along with our flexible attention mechanism allow for an expressive model that can provably capture shortest path distances and general families of graph propagations.
Theoretical results show that the RRWP initialization is strictly more expressive than shortest path distances when used in the GD-WL graph isomorphism test~\citep{zhang2023rethinking}, and a synthetic experiment shows that our flexible attention mechanism is indeed able to learn graph propagation matrices that other Graph Transformers are not as capable of learning.
Our GRIT model achieves state-of-the-art performance across a wide range of graph datasets, showing that data complexity can be improved for Graph Transformers without integrating local message-passing modules. 
Nonetheless, GRIT is not the final chapter for graph inductive biases in Transformers; future work can address limitations of our work, such as GRIT's $n^2$ scaling for updating pair representations, and lack of upper bounds on expressive power.

\section{Acknowledgments}
DL is supported by an NSF Graduate Fellowship.
CL is supported by Meta AI~\footnote{Meta has no relationships whatsoever with the other funding sponsors.}.
We would also like to thank the Royal Academy of Engineering and FiveAI.


\bibliographystyle{template/icml2023}
\bibliography{ref}

\newpage
\appendix
\onecolumn

\section{Model Architecture}

\subsection{Visualization of the Transformer Architecture of GRIT}\label{appendix:architecture}
\begin{figure*}[h]
    \centering
    \subfigure[The Overall Architecture of GRIT Transformers]{
    \centering
    \begin{minipage}[b]{.55\linewidth}
    \includegraphics[width=\textwidth]{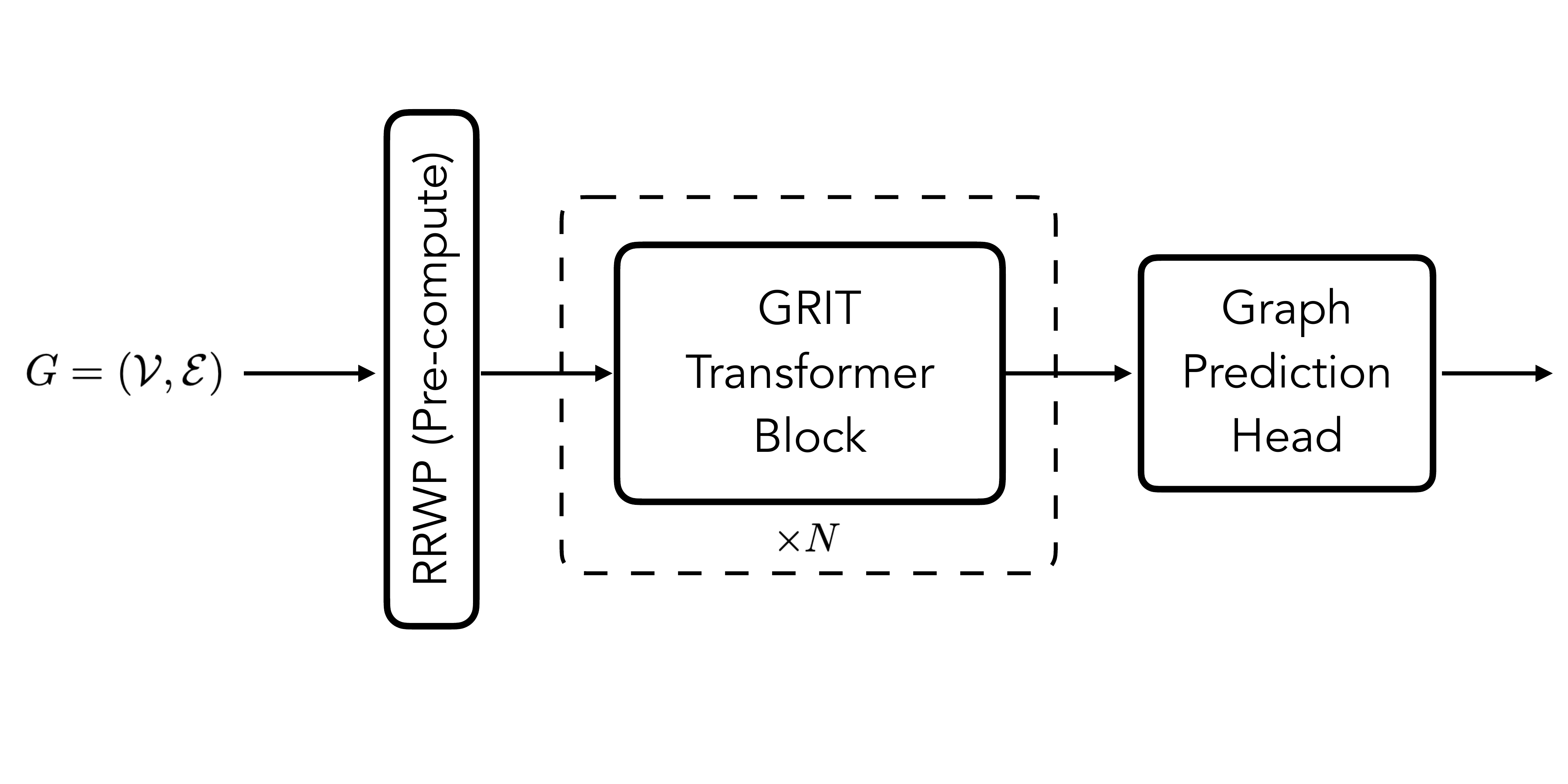}
    \end{minipage}
    }
    \subfigure[The Architecture of a GRIT Transformer Block]{
    \centering
    \begin{minipage}[b]{.25\linewidth}
    \includegraphics[width=\textwidth]{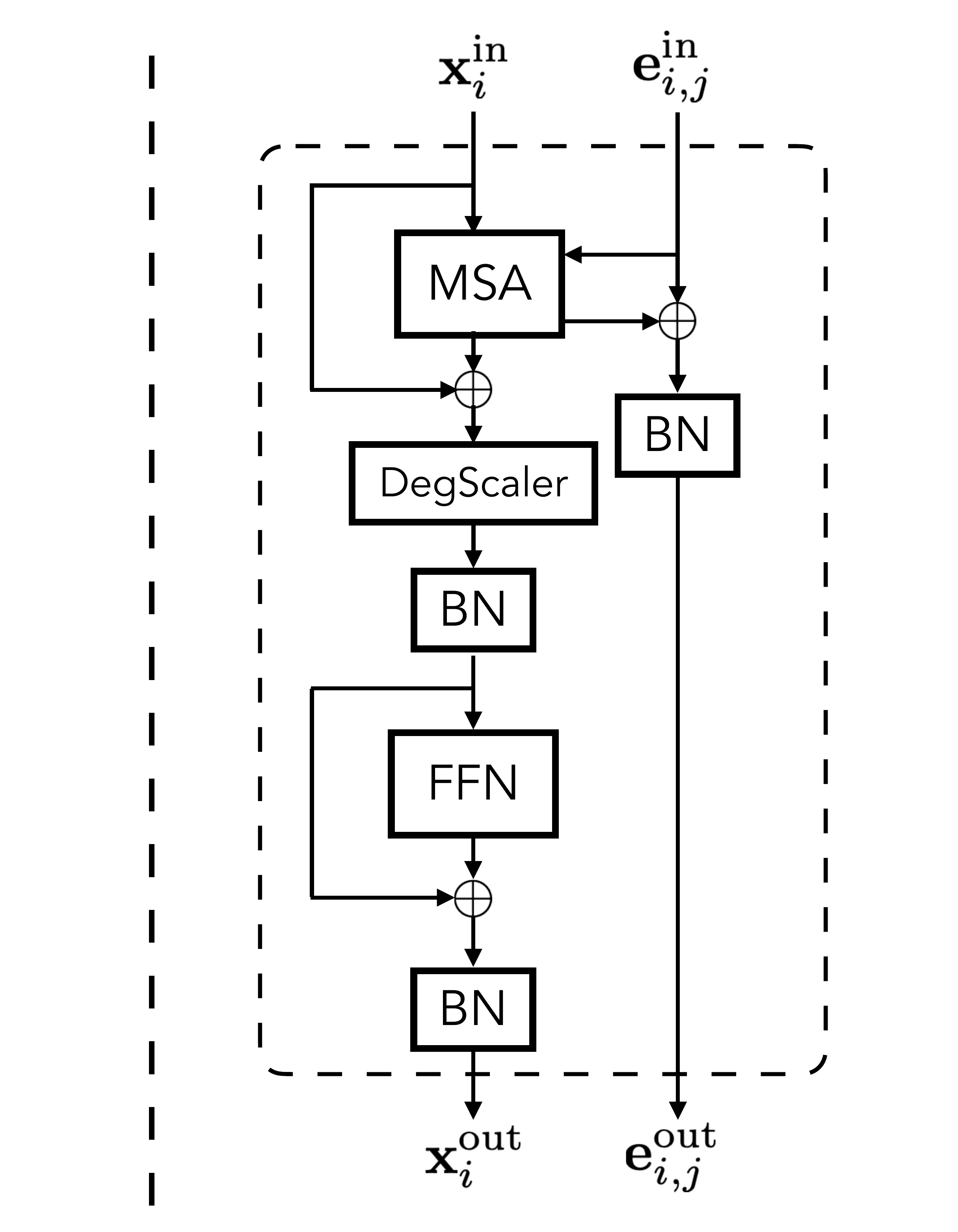}
    \end{minipage}
    }
    \caption{The Architecture of GRIT Transformers. (a) visualize the conceptural relationship between our proposed RRWP feature and the GRIT transformer. (b) shows the detailed design of GRIT transformer block.}
    \label{arch}
\end{figure*}

In order to put all the conceptual building blocks into one clear visualization, here we provide an overview of the GRIT transformer in Figure~\ref{arch}.
\vspace{-0.5cm}

\section{Experimental Details}
\label{appendix:experiment_details}

\subsection{Description of Datasets}

A summary of the statistics and characteristics of datasets is shown in Table.~\ref{tab:dataset}. The first five datasets are from \citet{dwivedi2020BenchmarkingGraphNeural}, the middle two are from \citet{dwivedi2022LongRangeGraph} and the last is from \citet{hu2021ogblsc}.
Readers are referred to \citet{rampasek2022RecipeGeneralPowerful} for more details about the datasets.


\begin{table}[h!]
    \centering
    \caption{Overview of the graph learning datasets involved in this work~\cite{dwivedi2020BenchmarkingGraphNeural, dwivedi2022LongRangeGraph, irwin2012ZINCFreeTool, hu2021ogblsc} .}
    \small
    \setlength{\tabcolsep}{1.6pt}
    \begin{tabular}{l|ccccccc}
    \toprule
       \textbf{Dataset} & \textbf{\# Graphs} & \textbf{Avg. \# nodes} & \textbf{Avg. \# edges}  &  \textbf{Directed} 
 & \textbf{Prediction level} & \textbf{Prediction task} & \textbf{Metric}\\
 \midrule
        ZINC(-full) & 12,000 (250,000) & 23.2 & 24.9 & No &  graph & regression & Mean Abs. Error \\
        MNIST &  70,000  &70.6  & 564.5 & Yes  & graph & 10-class classif. &  Accuracy \\
        CIFAR10 & 60,000 & 117.6 & 941.1 & Yes  & graph & 10-class classif. & Accuracy \\
        PATTERN & 14,000 & 118.9 & 3,039.3  & No & inductive node & binary classif. & Weighted Accuracy \\
        CLUSTER & 12,000 & 117.2 & 2,150.9 &  No & inductive node &  6-class classif. & Accuracy \\ 
        \midrule
        Peptides-func & 15,535 & 150.9 & 307.3 & No & graph & 10-task classif. &  Avg. Precision \\
        Peptides-struct &  15,535  & 150.9 & 307.3 & No & graph & 11-task regression &  Mean Abs. Error  \\
        \midrule
        PCQM4Mv2 & 3,746,620 & 14.1 & 14.6 & No & graph & regression & Mean Abs. Error \\
        \bottomrule
    \end{tabular}
    \label{tab:dataset}
\end{table}

\subsection{Dataset splits and random seed}
Our experiments are conducted on the standard train/validation/test splits of the evaluated benchmarks.
For each dataset, we execute 4 runs with different random seeds (0,1,2,3) and report the mean performance and standard deviation.

\subsection{Hyperparameters}

Due to the limited time and computational resources, we did not perform an exhaustive search or a grid search on the hyperparameters.
We mainly follow the hyperparameter setting of GraphGPS~\cite{rampasek2022RecipeGeneralPowerful} and make slight changes if the number of parameters does not fit in the commonly used parameter budgets.
For the benchmarks from \citet{dwivedi2020BenchmarkingGraphNeural, dwivedi2022LongRangeGraph}, we follow the most commonly used parameter budgets: up to 500k parameters for ZINC, PATTERN, CLUSTER, Peptides-func and Peptides-struct; and ~100k parameters for MNIST and CIFAR10.

The final hyperparameters are presented in Tables.~\ref{tab:bmgnn_hparam} and Tables.~\ref{tab:lrgb_hparam}.

\begin{table}[h!]
    \centering
    \caption{Hyperparameters for five datasets from BenchmarkingGNNs \cite{dwivedi2020BenchmarkingGraphNeural} and ZINC-full~\cite{irwin2012ZINCFreeTool}
    }
    \label{tab:bmgnn_hparam}
\begin{tabular}{lccccc}
\toprule
Hyperparameter & ZINC/ZINC-full & MNIST & CIFAR10 & PATTERN & CLUSTER \\
\midrule
\# Transformer Layers & 10 & 3 & 3 & 10 & 16 \\
Hidden dim & 64 & 52 & 52 & 64 & 48 \\
\# Heads & 8 & 4 & 4 & 8 & 8 \\
Dropout & 0 & 0 & 0 & 0 & $0.01$ \\
Attention dropout & $0.2$ & $0.5$ & $0.5$ & $0.2$ & $0.5$ \\
Graph pooling & sum & mean & mean & $-$ & $-$ \\
\midrule
PE dim (RW-steps) & 21 & 18 & 18 & 21 & 32 \\
PE encoder & linear & linear & linear & linear & linear \\
\midrule
Batch size & 32/256 & 16 & 16 & 32 & 16 \\
Learning Rate & $0.001$ & $0.001$ & $0.001$ & $0.0005$ & $0.0005$ \\
\# Epochs & 2000 & 200 & 200 & 100 & 100 \\
\# Warmup epochs & 50 & 5 & 5 & 5 & 5 \\
Weight decay & $1 \mathrm{e}-5$ & $1 \mathrm{e}-5$ & $1 \mathrm{e}-5$ & $1 \mathrm{e}-5$ & $1 \mathrm{e}-5$ \\
\midrule
\# Parameters & 473,473 & 102,138 & 99486 & 477,953 & 432,206 \\
\bottomrule
\end{tabular}
\end{table}

\begin{table}[ht]
    \centering
        \caption{Hyperparameters for two datasets from the Long-range Graph Benchmark~\cite{dwivedi2022LongRangeGraph} and PCQM4Mv2~\cite{hu2021ogblsc}}
    \label{tab:lrgb_hparam}
    \begin{tabular}{lccc}
\midrule Hyperparameter & Peptides-func & Peptides-struct & PCQM4Mv2 \\
\midrule \# Transformer Layers & 4 & 4  & 16 \\
Hidden dim &  96 & 96  & 256 \\
\# Heads & 4 & 8  & 8 \\
Dropout & 0 & 0 & 0.1\\
Attention dropout &  0.5 & 0.5 & 0.1 \\
Graph pooling &  mean & mean & mean \\
\midrule
PE dim (walk-step) & 17 & 24  & 16 \\
PE encoder & linear & linear & linear \\
\midrule
Batch size & 32 & 32 & 256 \\
Learning Rate &  0.0003 & 0.0003 & 0.0002\\
\# Epochs & 200 & 200 & 150 \\
\# Warmup epochs &  5 & 5 & 10 \\
Weight decay &  0 & 0  & 0\\
\midrule
\# Parameters & 443,338 & 438,827 & 15.3M \\
\bottomrule
\end{tabular}
\end{table}

\subsection{Significance Test}

We conduct a two-sample one-tailed T-test to compare the results of our method with the second-best model on each dataset.
The baselines' results are  taken from \cite{rampasek2022RecipeGeneralPowerful}, with 10 runs for datasets from \cite{dwivedi2020BenchmarkingGraphNeural} and 4 runs for datasets from  \cite{dwivedi2022LongRangeGraph}.

The statistical tests are conducted using the tools available at \url{https://www.statskingdom.com/140MeanT2eq.html}.

\subsection{Visualization for the Synthetic Experiment}\label{appendix:synthetic}

\begin{figure*}[h!]
    \centering
    \includegraphics[width=1\textwidth]{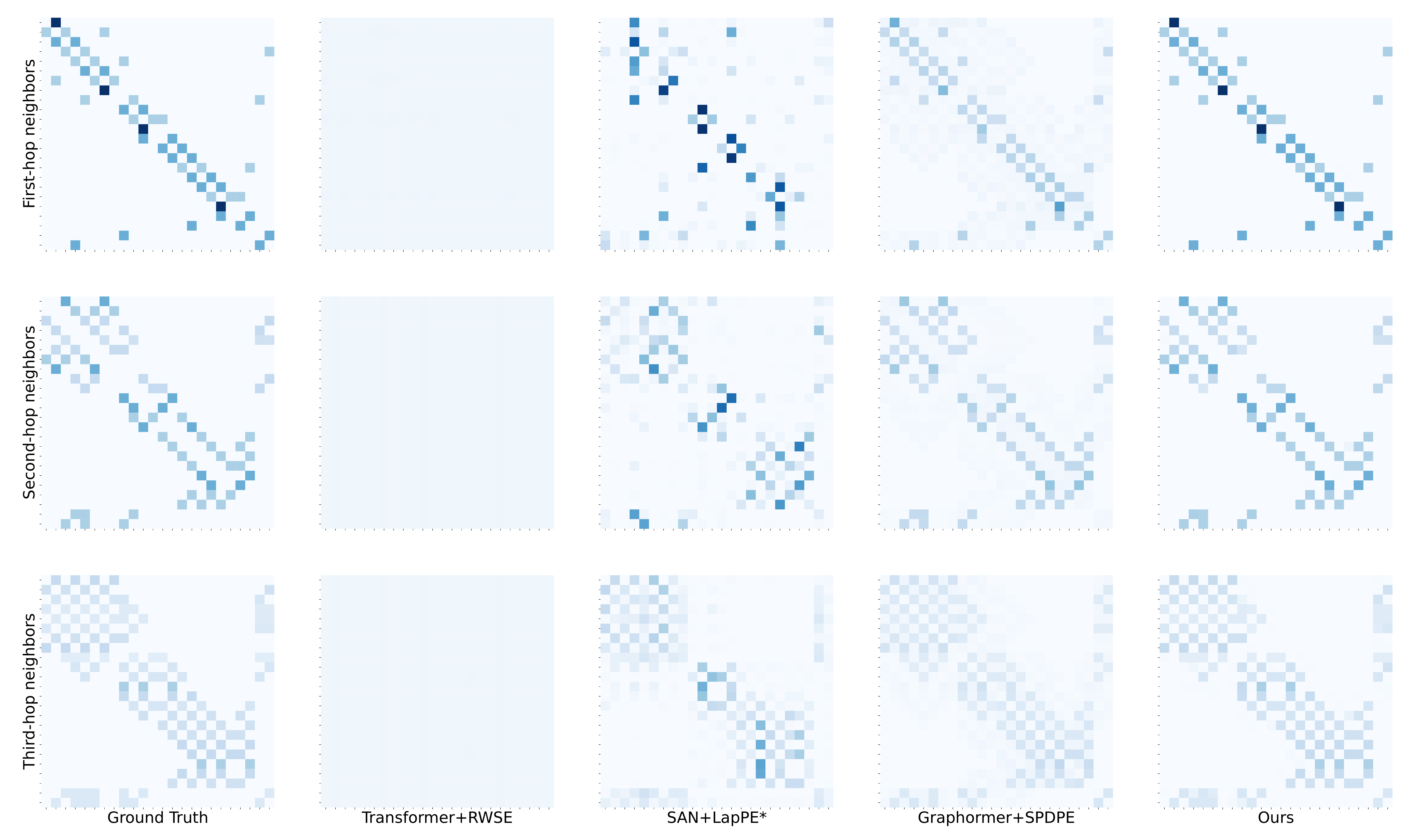}
    
    \caption{Visualization of learned attention scores for the synthetic experiment on learning to attend to ($k=1,2,3$)-hop neighbors. Our GRIT attention mechanism (far right) is the most successful at matching both the sparsity pattern and attention magnitudes of the target (far left).
    }
    \label{fig:visual_attn}
\end{figure*}

In Figure~\ref{fig:visual_attn}, we visualize the learned attention scores for a single graph from our synthetic experiments in Section~\ref{sec:attn2mp}. Recall that the goal of the synthetic experiments was to test the ability of different attention mechanisms and positional encoding schemes to learn to attend to $k$-hop neighborhoods, using only a single layer of attention and only the PE as input.


Transformer+RWSE struggles to learn to attend to $k$-hop neighbors.
Graphormer+SPDPE can successfully attend to most nodes in $k$-hop neighborhoods; however, many other nodes are still assigned small attention scores, so the fraction of attention focused on the neighborhoods is diminished. 
SAN+LapPE* is better at capturing the scales of the target attention scores, but it assigns high attention scores to multiple nodes outside the targeted neighborhoods.
In contrast, our GRIT attention mechanism can successfully attend to the nodes in $k$-hop neighborhoods.
We also provide the result table of the synthetic experiment with $R^2$ metric, shown in Table.~\ref{tab:attn2mpR2}.

\begin{table}[ht]
    \centering
    \setlength{\tabcolsep}{2pt}
    \caption{Synthetic Experiment on Learning to Attend $K$-hop Neighborhoods}
    \label{tab:attn2mpR2}
    {\small
\begin{tabular}{l|lll}
    \toprule
        $R^2$ $\leq 1$  & 1-hop & 2-hop & 3-hop  \\ \midrule
        MeanPool (baseline) & $ 0 \pm 0$ & $0 \pm 0$ & $0 \pm 0$  \\ \midrule
        Transf.+RWSE & $ 0 \pm 0$ & $0 \pm 0$ & $0 \pm 0$  \\ 
        SAN+LapPE & $0.32 \pm 0.20$ & $0.31 \pm 0.24 $ & $0.52 \pm 0.21$  \\ 
        EGT+SVDPE & $ 0.55 \pm 0.21 $ & $ 0.28\pm 0.25$ & $0.26 \pm 0.28 $  \\ 
        Graphormer+SPDPE & $ 0.67 \pm 0.09 $ & $0.77 \pm 0.08$ & $0.80\pm 0.06$  \\ \midrule
        GRIT (Ours) & $\first{\mathbf{0.999 \pm 0.001}}$ & $\first{\mathbf{0.998 \pm 0.004}}$ & $\first{\mathbf{0.961 \pm 0.035}}$ \\ \bottomrule
    \end{tabular}
    }
\end{table}

\subsection{Asymptotic Complexity, Runtime and GPU Memory}\label{appendix:runtime}

The asymptotic complexities of RRWP and GRIT’s attention mechanism are $O(K|\mathcal{V}||\mathcal{E}|)$
and $O(|\mathcal{V}|^2)$
respectively, where 
$K$ is the number of hops of RRWP, 
$|\mathcal{E}|$ is the number of edges and 
$|\mathcal{V}|$ is the number of nodes,
matching the asymptotic complexity of most Graph Transformers~\cite{kreuzer2021RethinkingGraphTransformers, rampasek2022RecipeGeneralPowerful, hussain2022GlobalSelfAttentionReplacement, ying2021TransformersReallyPerform}.
. 
 
We provide the runtime and GPU memory consumption of GRIT and other baselines on ZINC as a reference (Table.~\ref{tab:runtime_memory}).
The runtime is given by the pipeline of GraphGPS~\cite{rampasek2022RecipeGeneralPowerful}, and the memory usage is counted by the NVIDIA System Management Interface (nvidia-smi).

\begin{table}[!ht]
    \centering
    \caption{Runtime and GPU memory for SAN~\cite{kreuzer2021RethinkingGraphTransformers}, GraphGPS~\cite{rampasek2022RecipeGeneralPowerful} and GRIT (Ours) on ZINC with batch size $32$. The timing is conducted on a single NVIDIA V100 GPU and 20 threads of Intel(R) Xeon(R) GOld 6140 CPU @ 2.30GH.}
    \begin{tabular}{l|lll}
    \toprule
        ZINC & SAN & GraphGPS & GRIT (Ours) \\ \midrule
        MAE $\downarrow$ & $0.139 \pm 0.006$ & $0.070 \pm 0.004$ & $0.059 \pm 0.002$ \\ 
        PE Precompute-time & 10 sec & 11 sec & 11 sec \\ 
        GPU Memory & 2291 MB & 1101 MB & 1865 MB \\
        Training time & 57.9 sec/epoch & 24.3 sec/epoch & 29.4 sec/epoch \\ \bottomrule
    \end{tabular}
    \label{tab:runtime_memory}
\end{table}


\begin{figure*}[ht]
\newcommand{\wth}{.19}

\centering
{
\hspace{6pt} \includegraphics[width=\wth\textwidth]{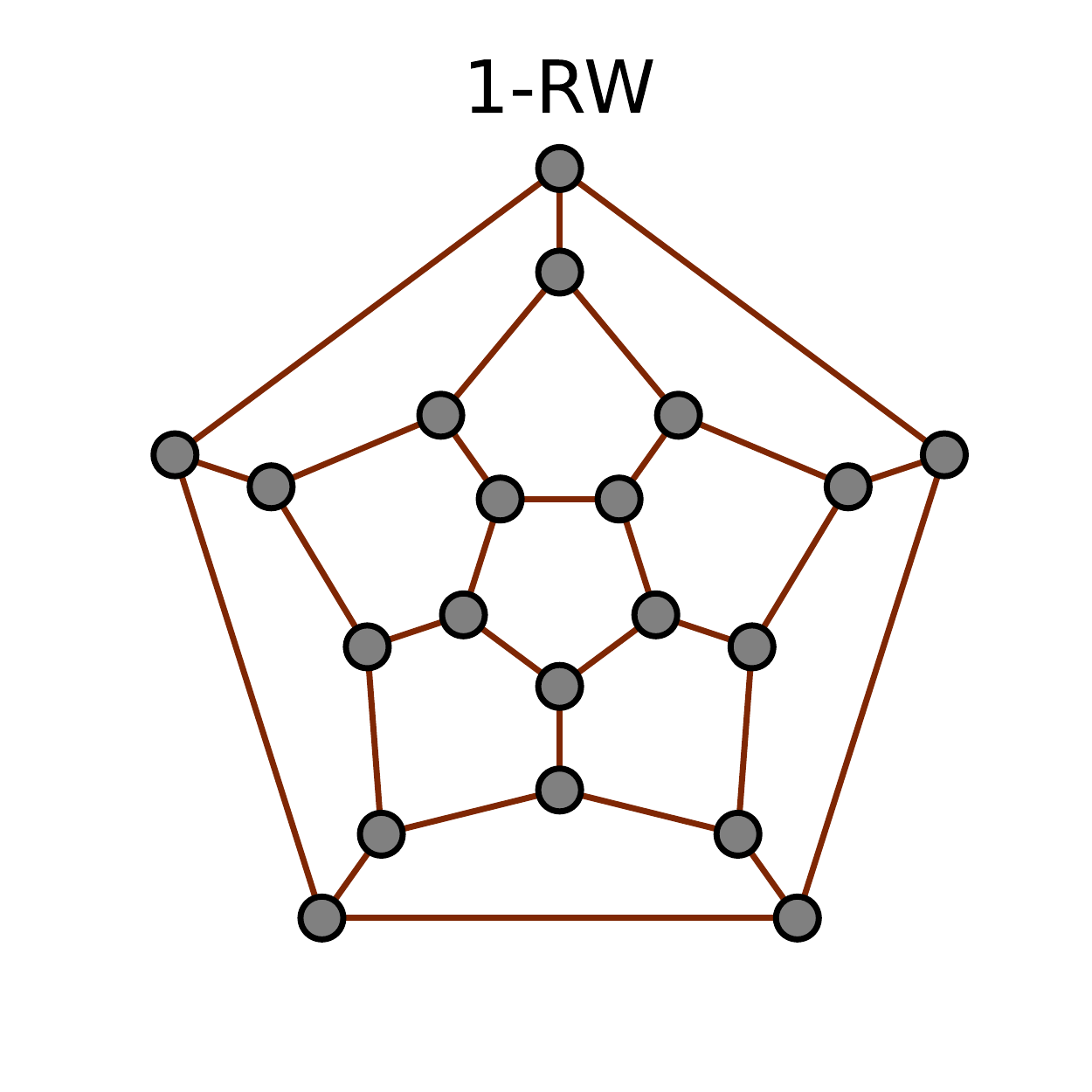}
\includegraphics[width=\wth\textwidth]{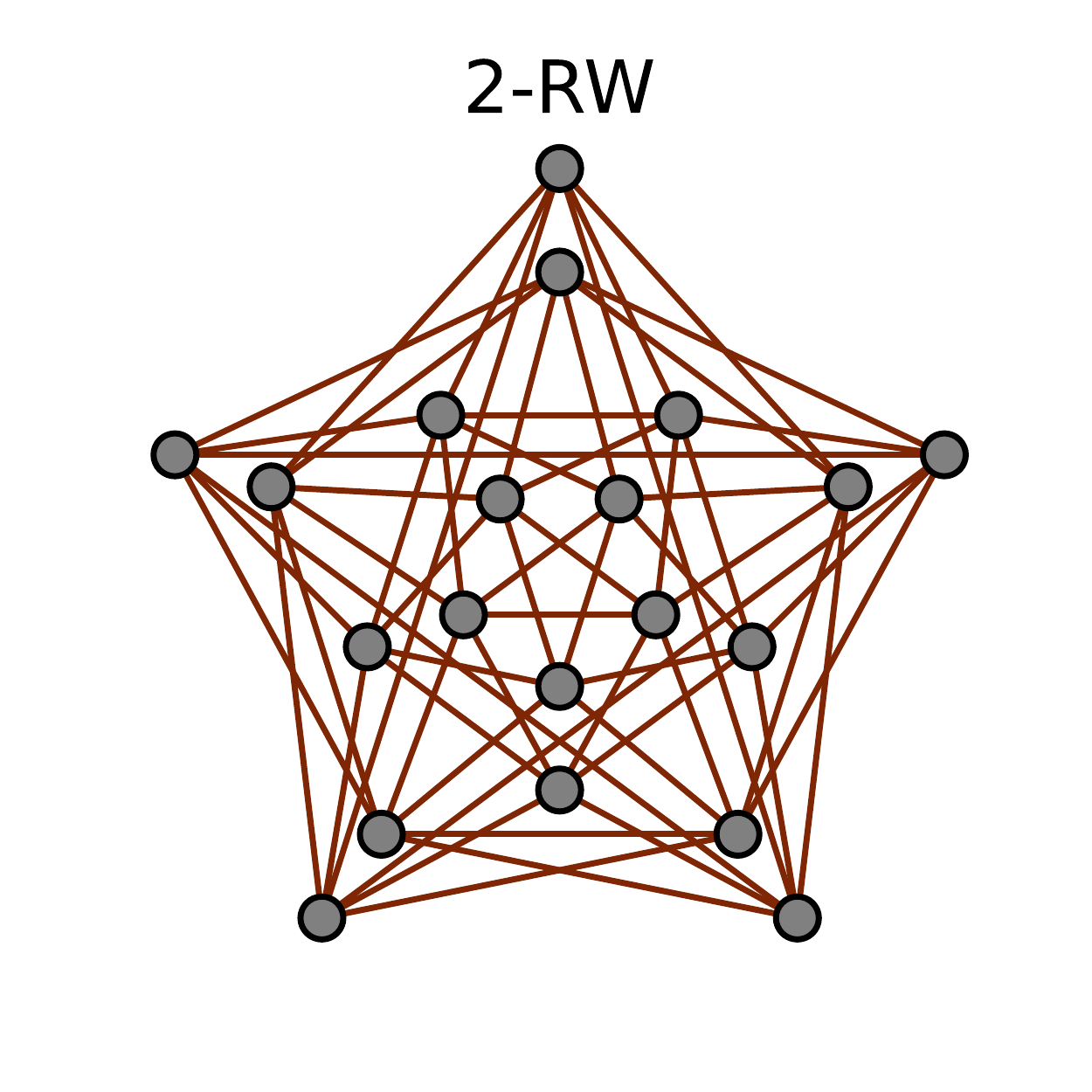}
\includegraphics[width=\wth\textwidth]{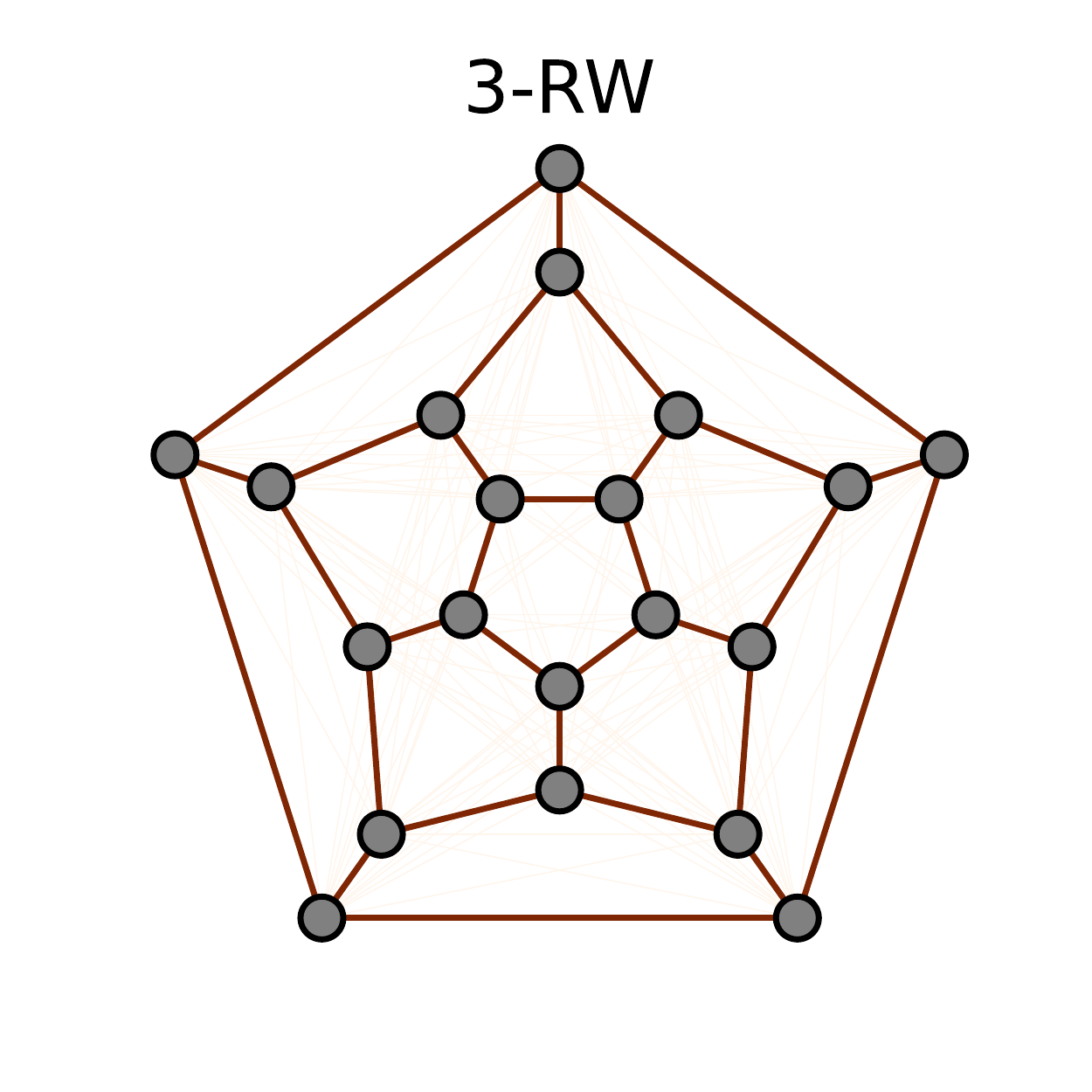}
\includegraphics[width=\wth\textwidth]{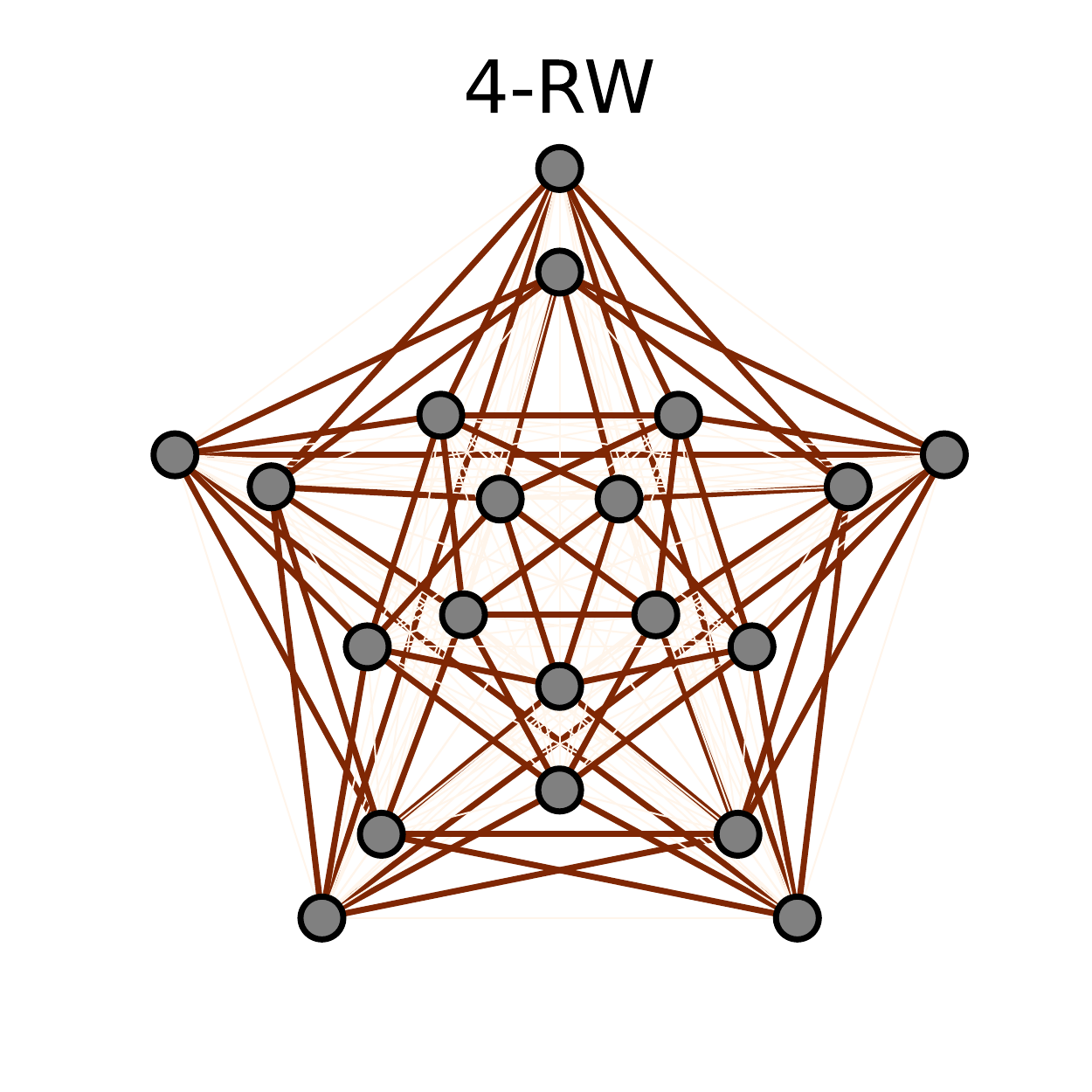}
\includegraphics[width=\wth\textwidth]{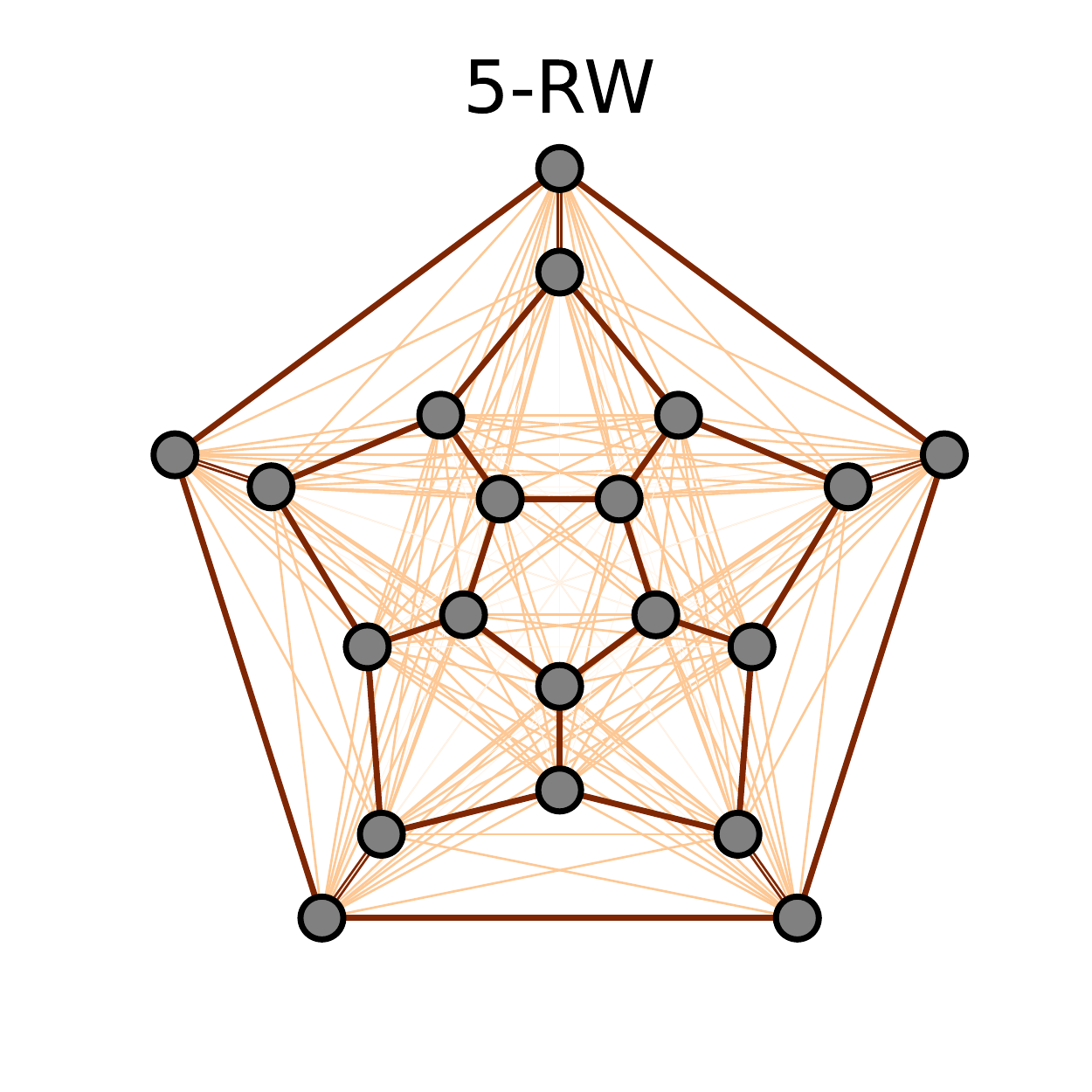}\\
}
\centering
{
\hspace{8.5pt} \includegraphics[width=\wth\textwidth]{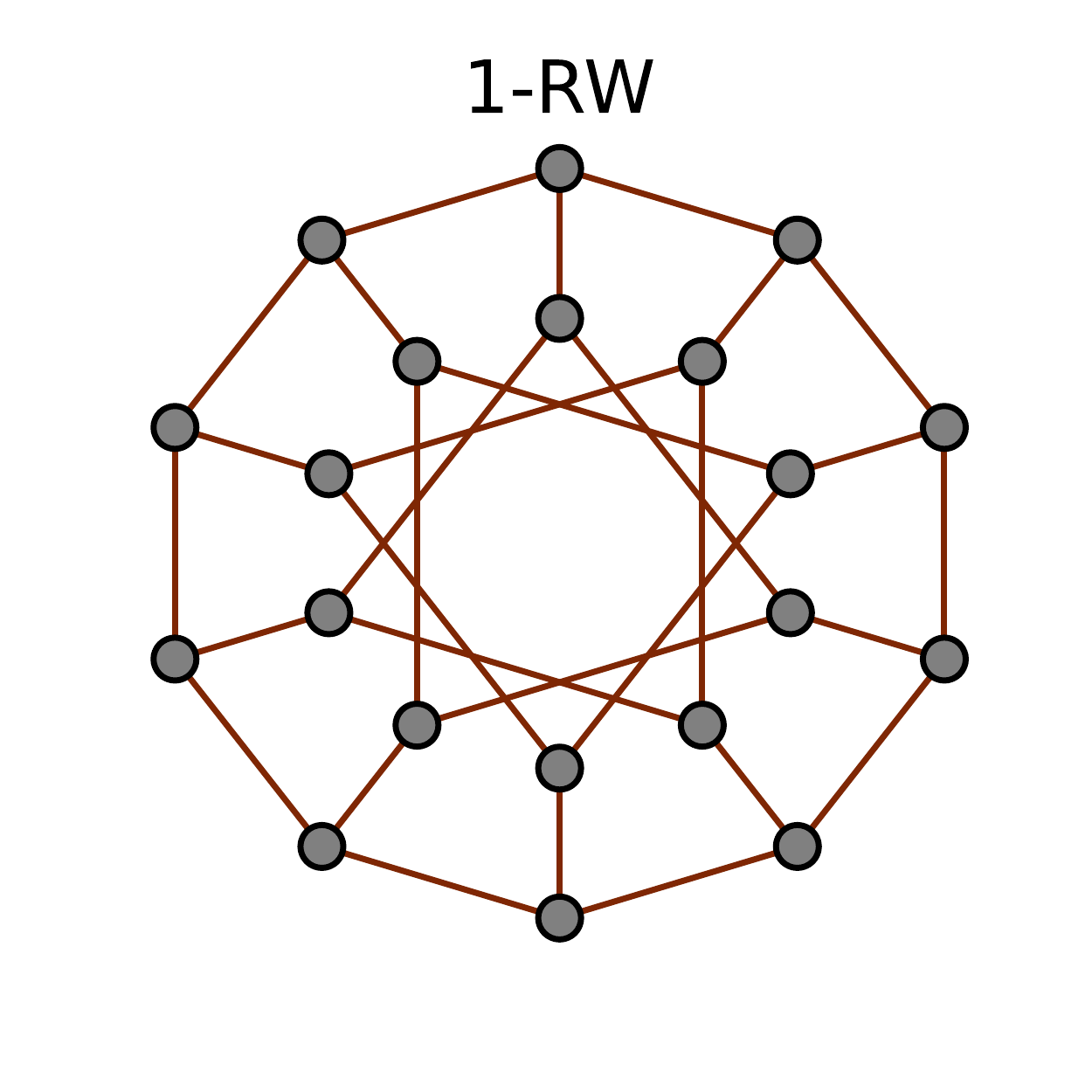}
\includegraphics[width=\wth\textwidth]{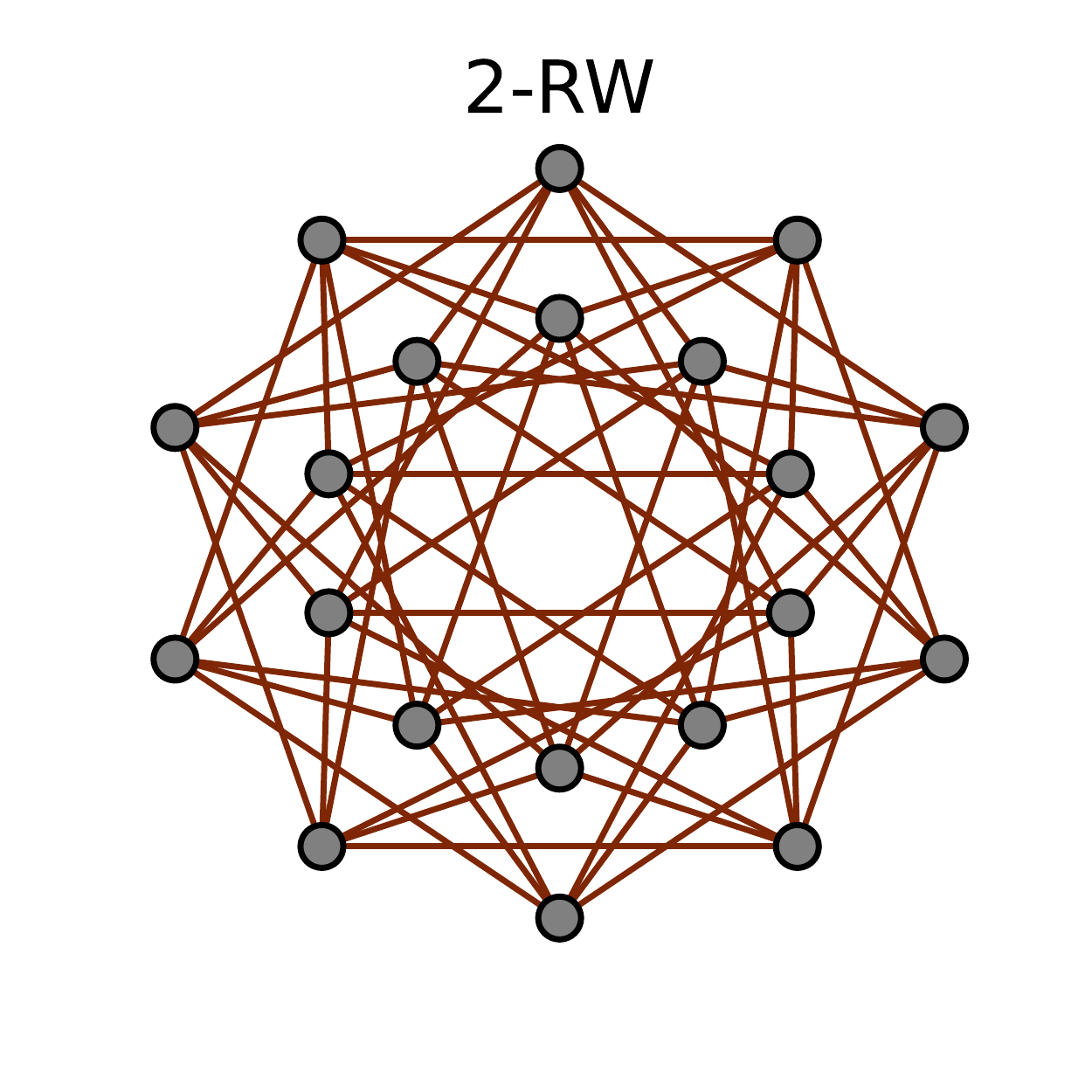}
\includegraphics[width=\wth\textwidth]{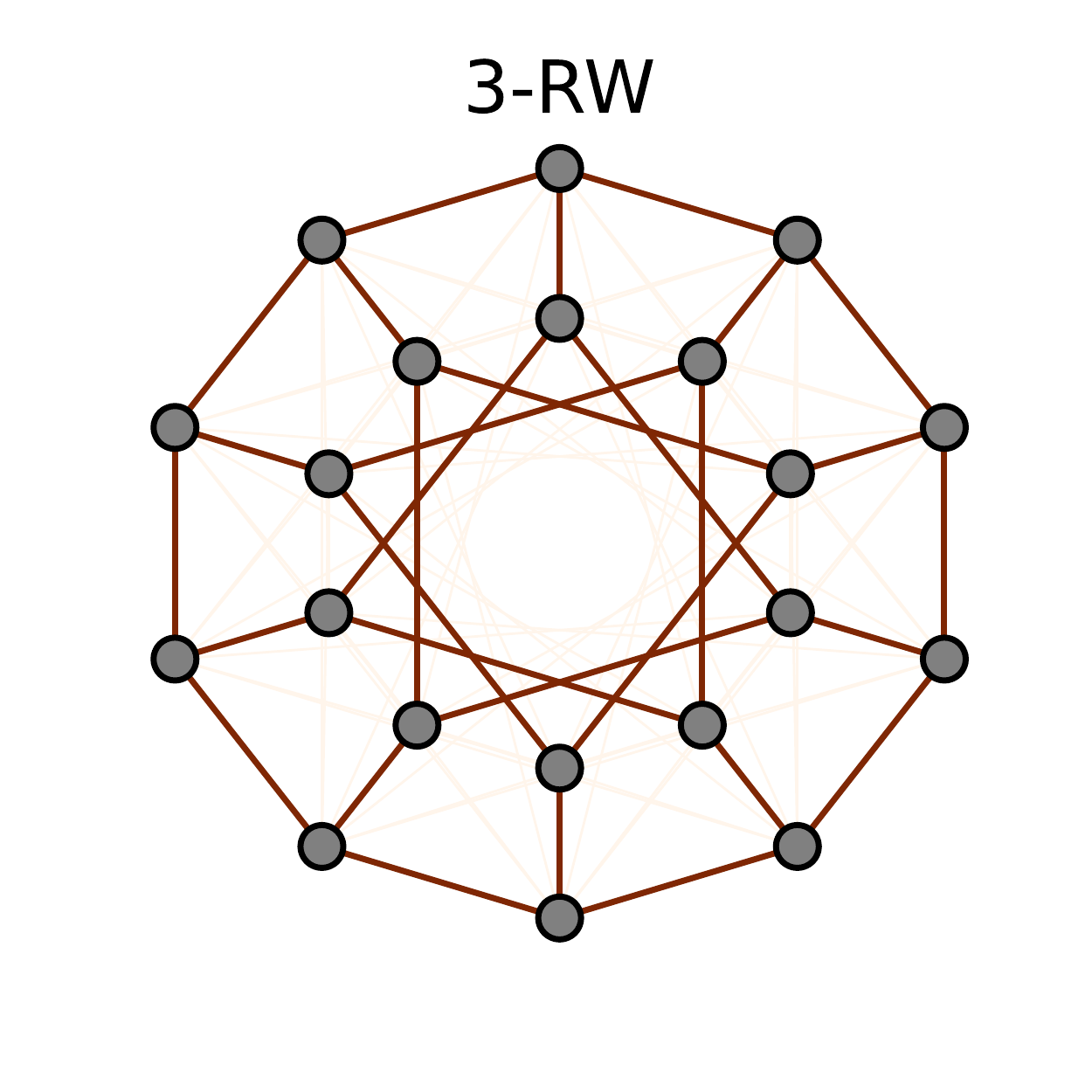}
\includegraphics[width=\wth\textwidth]{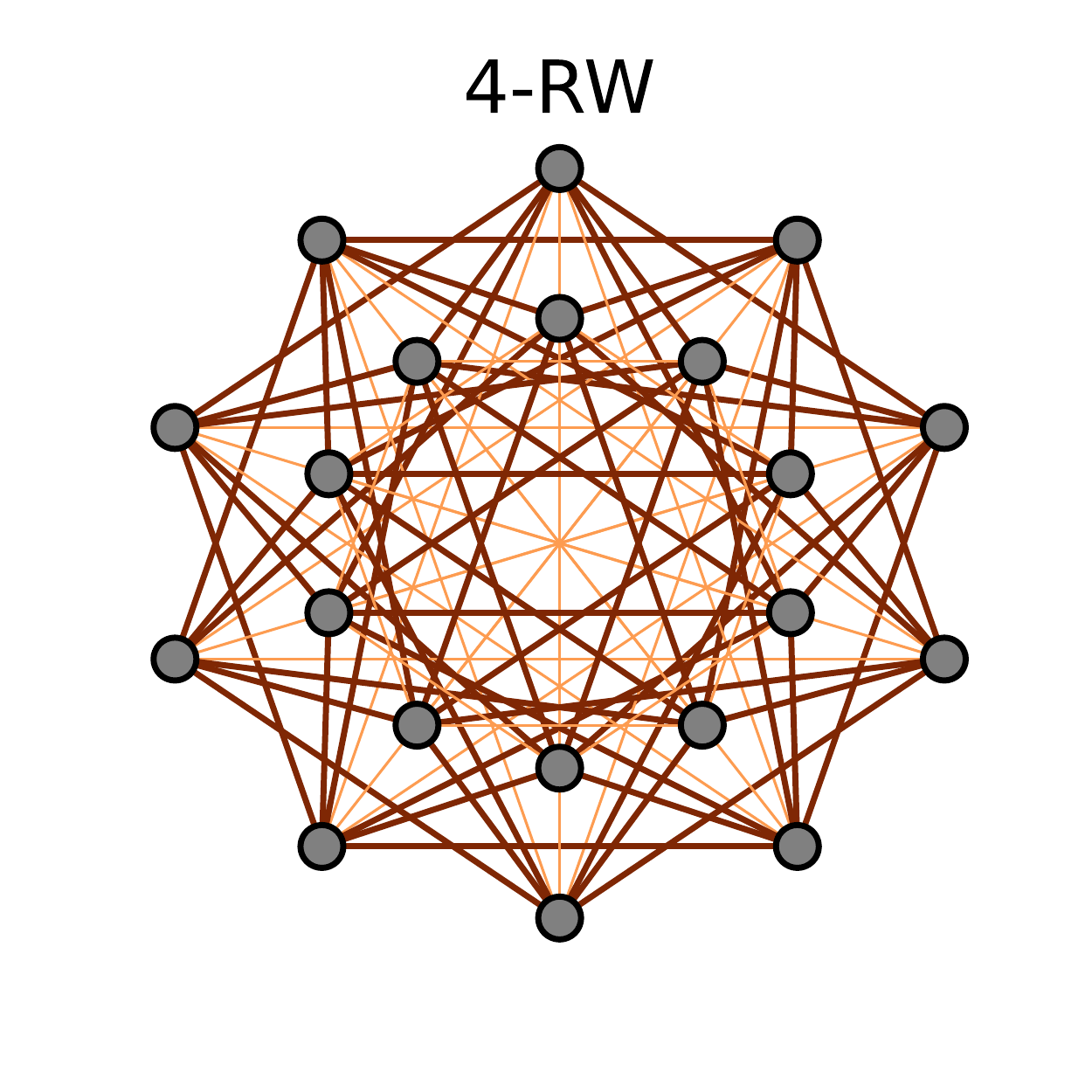}
\includegraphics[width=\wth\textwidth]{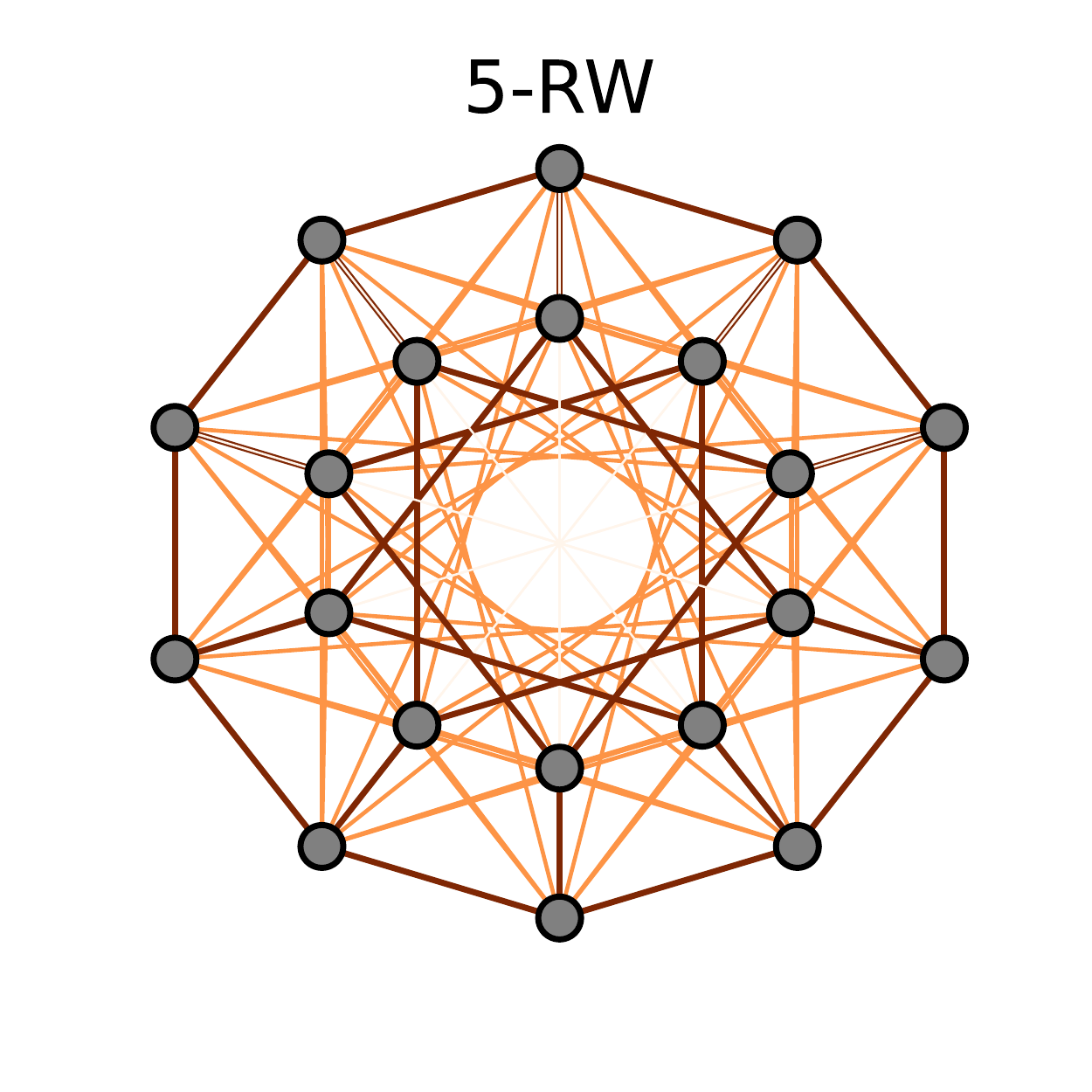}
}
\caption{(Top row) RRWP for the Dodecahedron graph. (Bottom row) RRWP for the  Desargues graph. This pair of non-isomorphic graphs cannot be distinguished by GD-WL with shortest path distances, but can be distinguished by GD-WL with our RRWP positional encoding.}
\label{fig:rrwp_distinguish}
\end{figure*}

\section{Proofs of Theoretical Results}

\subsection{GD-WL}\label{appendix:gdwl_proof}

\begin{proof}[Proof of Proposition~\ref{prop:rrwp_gdwl}]
    First, we show that GD-WL with RRWP is at least as strong as GD-WL with shortest path distances. Then we give an example of two graphs that GD-WL with shortest path distances cannot distinguish, yet GD-WL with RRWP can.
    
    Let $d_G^{\mathrm{RW}}(v,u) \in \RR^n$ be the relative random walk encoding with $K = n$, and $d_G^{\mathrm{SPD}} \in \RR$ be the shortest path distance encoding. Then note that 
    \begin{equation}
     \min \{i :\  d_G^{\mathrm{RW}}(v,u)_i \neq 0 \}   = d_G^{\mathrm{SPD}}(v,u) \},   
    \end{equation}
    where the $\min$ takes the value $\infty$ if $d_G^{\mathrm{RW}}(v,u)_i = 0$ for each $i$.
     Thus, $d_G^{\mathrm{SPD}}$ is a function of $d_G^{\mathrm{RW}}$, and hence $d_G^{\mathrm{RW}}$ refines $d_G^{\mathrm{SPD}}$. We finish by using Lemma 2 of \citet{bevilacqua2022EquivariantSubgraphAggregation}, which says that taking multisets of colors preserves refinement. This shows that GD-WL with RRWP is at least as strong as GD-WL with SPD.

    To show that GD-WL with RRWP is strictly stronger, we present a pair of non-isomorphic graphs that it can distinguish but GD-WL with SPD cannot: the Desargues graph and the Dodecahedral graph, which are plotted in Figure~\ref{fig:rrwp_distinguish}. \citet{zhang2023rethinking} note that GD-WL with shortest path distances cannot distinguish these graphs. However, GD-WL with our RRWP positional encoding can. This can be seen from the 5-hop random walk probability distributions --- there exists at least one walk of exactly length 5 between any two nodes of the Dodecahedral graph, but there is no exactly length 5 walk between many pairs of nodes of the Desargues graph.
\end{proof}

\subsection{RRWP and MLP}\label{appendix:rrwp_spd_prop}

\begin{proof}[Proof of Proposition~\ref{prop:mlp_rrwp}]
(a) First, we construct an $\mlp$ such that $\mlp(\P)_{i,j} \approx \spd_{K-1}(i,j)$, in which $\spd_{K-1}(i,j)$ takes the value of the length of the shortest path between nodes $i$ and $j$ if $i$ and $j$ are no more than $K-1$ hops away from each other, and takes the value of $K$ otherwise.

We build the $\mlp$ to approximate a composition of several continuous functions $f_3 \circ f_2 \circ f_1$. Let $\P(\A)$ be the RRWP associated to the adjacency matrix $\A$, i.e. $\P(\A) = [\mathbf{I}, \D^{-1}\A, \ldots, (\D^{-1}\A)^{K-1}]$.
We define $L$ to be a lower bound on the smallest nonzero entry of $\P(\A)$, across all $\A \in \mathbb{G}_n$. In particular, we let
\begin{equation}
    L = \min_{\A \in \mathbb{G}_n} \  \min_{i,j,t : \P(\A)_{ijt} > 0} \P(\A)_{ijt}.
\end{equation}
Since the minimizations are over finitely many positive entries, we have that $L > 0$. Thus, there exists a continuous function $f_1: \RR^{K} \to \RR^{K}$ such that $f_1(x)_i = 0$ if $x_i \leq 0$ and 1 if $x_i \geq L$. Since for $t \in \{0, \ldots, K-1\}$ $\P_{ijt} \geq L$ if and only if node $i$ is reachable to $j$ in $t$ hops, we have that
\begin{equation}
    f_1(\P_{i, j, :})_t = \begin{cases}
        1 & \text{if $i$ can reach $j$ in $t$ hops}\\
        0 & \text{else}
    \end{cases}.
\end{equation}
Next, we define $f_2: \RR^{K} \to \RR^{K}$ as $f_2(x)_{t} = \max_{t' \leq t} x_{t'}$. Then we have
\begin{equation}
     (f_2 \circ f_1(\P_{i, j, :}))_t = \begin{cases}
        1 & \text{if $\spd(i,j) \leq t$}\\
        0 & \text{else}
    \end{cases}.
\end{equation}
Finally, we let $f_3: \RR^{K} \to \RR$ be defined by $f_3(x) = n - \sum_{t=1}^{K-1} x_t$. The full composition then gives
\begin{equation}
    f_3 \circ f_2 \circ f_1(\P_{i, j, :}) = \begin{cases}
        \spd(i,j) & \text{if $\spd(i,j) \leq K-1$}\\
        n & \text{else}
    \end{cases}.
\end{equation}
Now, note that $f_3 \circ f_2 \circ f_1$ is continuous, so by standard universal approximation results~\citep{hornik1989multilayer} we can approximate it with an $\mlp$ on the compact set $\mathbb{G}_n$ to an arbitrary accuracy $\epsilon > 0$. This finishes the proof.

(b) Now, fix $\theta_0, \ldots, \theta_{K-1} \in \RR$. We will construct an $\mlp$ such that $\mlp(\P) \approx \sum_{k=0}^{K-1} \theta_k (\D^{-1}\A)^k$. Note that this target function can be written as
\begin{equation}
    \sum_{k=0}^{K-1} \theta_k \rw^k = \sum_{k=0}^{K-1} \theta_k \P_{:, :, k}.
\end{equation}
Hence, we let $f_1: \RR^K \to \RR^k$ be the continuous function that scales a vector $x \in \RR^K$ elementwise by $\btheta = [\theta_0, \ldots, \theta_{K-1}]$: $f_1(x) = x \odot \btheta$. Then let $f_2: \RR^K \to \RR$ take the sum: $f_2(x) = \sum_{k=0}^{K-1} x_k$. It is easy to see that
\begin{equation}
    f_2 \circ f_1(\P_{i,j}) = \sum_{k=0}^{K-1} \theta_k \P_{i,j,k},
\end{equation}
so letting the $\mlp$ approximate the continuous function $f_2 \circ f_1$, we are done.

(c) Finally, we will construct an $\mlp$ such that $\mlp(\P) \approx \theta_0 \mathbf{I} + \theta_1 \A$. Note that this target function is equal to $\theta_0 \P_{:, :, 0} + \theta_1 \A$.

To get the adjacency $\A$, we use the function $f_1: \RR^K \to \RR^K$ as used in the proof of part (a), which rounds $f_1(x)_i = 0$ if $x_i \leq 0$ and $1$ if $x_i \geq L$. Then $f_1(\P_{i,j})_1 = \A_{i,j}$. Then we take $f_2: \RR^K \to \RR^K$ to scale the first two entries by $\theta_0$ and $\theta_1$ respectively (similarly to $f_1$ in the proof of part (b)), and take $f_3: \RR^K \to \RR$ to sum across the first two slices: $f_3(x) = x_0 + x_1$. Then we have that
\begin{equation}
    f_3 \circ f_2 \circ f_1(\P_{i,j}) = \theta_0 \mathbf{I} + \theta_1 \A
\end{equation}
as desired. Choosing an $\mlp$ that approximate this continuous function $f_3 \circ f_2 \circ f_1$, we are done.
\end{proof}

\subsection{LayerNorm on Nodes Removes the Degree Information from Sum-Aggregators and/or Degree Scalers}\label{appendix:layernorm_degree}

Normalization layers are essential for deep neural networks, especially Transformers~\citep{vaswani2017AttentionAllYou}.
Graph Transformers usually use BatchNorm~\cite{ioffe2015BatchNormalizationAccelerating}, following most MPNNs, or LayerNorm~\citep{ba2016LayerNormalization}, following Transformers in other domains~\citep{vaswani2017AttentionAllYou, dosovitskiy2020ImageWorth16x16}.
Here, we show that a LayerNorm removes degree information, which motivates our choice of BatchNorm in our Transformer.

\begin{proof}[Proof of Proposition~\ref{prop:ln_degree}]
As noted by~\citet{corso2020PrincipalNeighbourhoodAggregation}, 
the output representation for a node $i$ from a sum-aggregator can be viewed as
$\mathbf{x}^\text{sum}_i =  d_i \cdot \mathbf{x}^\text{mean}_i$ where $d_i \in \mathbb{R}$ is the degree of node $i$ and $\mathbf{x}^\text{mean}_i=[x_{i1}, \dots x_{iF}]^\top \in \mathbb{R}^F$ is the node representation from a mean-aggregator.
The layer normalization statistics for a node $i$ over all hidden units are computed as follows:
\begin{equation}
    \begin{aligned}
    & \mu^\text{sum}_i = \frac{1}{F} \sum_{j=1}^F x^\text{sum}_{ij} = 
    \frac{1}{F} \sum_{j=1}^F d_i \cdot x^\text{mean}_{ij} 
    = \frac{d_i}{F} \sum_{j=1}^F x^\text{mean}_{ij} 
    = d_i \cdot \mu^\text{mean}_i \\
    &\sigma^\text{sum}_i =
    \sqrt{\frac{1}{F}\sum_{j=1}^F (x^\text{sum}_{ij}-\mu^\text{sum})^2}
    = 
    \sqrt{\frac{d_i^2}{F}\sum_{j=1}^F (x^\text{mean}_{ij}-\mu^\text{mean})^2} = d_i \cdot \sigma_i^\text{mean}
    \end{aligned}
\end{equation}

Therefore, regardless of the elementwise affine transforms shared by all nodes, each element of 
the normalized representation 
\begin{equation}
    \begin{aligned}
        \tilde{x}_{ij}^\text{sum} = 
        \frac{({x}_{ij}^\text{sum} - \mu_i^\text{sum})}{\sigma^\text{sum}_i} 
        = 
        \frac{(d_i \cdot {x}_{ij}^\text{mean} - d_i\cdot \mu_i^\text{mean})}{d_i \cdot \sigma^\text{mean}_i}
        = 
        \frac{( {x}_{ij}^\text{mean} - \mu_i^\text{mean})}{\sigma^\text{mean}_i} = \tilde{x}_{ij}^\text{mean},
        \quad \forall i \in \mathcal{V}, \forall j=1,\dots, F, 
    \end{aligned}
\end{equation}
will be the same for both sum-aggregation and mean-aggregation. 

The same conclusion can be seen for degree scalers, by simply changing $d_i$ to $f(d_i)$ in the proof, where $f: \mathbb{R} \to \mathbb{R}_{> 0}$.

\end{proof}


\end{document}